\def\year{2019}\relax
\documentclass[letterpaper]{article} 
\usepackage{aaai19}  
\usepackage{times}  
\usepackage{helvet}  
\usepackage{courier}  
\usepackage{url}  
\usepackage{graphicx}  
\usepackage{algorithm}
\usepackage{algorithmic}
\usepackage{amsmath,bm}
\usepackage{amssymb, multirow, paralist}
\usepackage{amsthm}
\usepackage{bbm}
\usepackage{caption}

\def \R {\mathbb{R}}

\def \P {\mathcal{P}}
\def \x {\mathbf{x}}
\def \OO {\mathcal{O}}

\def \p {\mathbf{p}}
\def \q {\mathbf{q}}

\def \D {\mathcal{D}}

\def \LL {\mathcal{L}}

\def \f {\mathbf{f}}
\def \F {\mathcal{F}}

\newtheorem{thm}{Theorem}
\newtheorem{prop}{Proposition}
\newtheorem{cor}{Corollary}
\newtheorem{lemma}{Lemma}
\newtheorem{definition}{Definition}

\begin{document}
%
\title{Robust Optimization over Multiple Domains}
\author{Qi Qian\quad Shenghuo Zhu\quad Jiasheng Tang\quad Rong Jin\quad Baigui Sun\quad Hao Li\\
Alibaba Group, Bellevue, WA, 98004, USA\\
\{qi.qian, shenghuo.zhu, jiasheng.tjs, jinrong.jr, baigui.sbg, lihao.lh\}@alibaba-inc.com\\
}
\maketitle
\begin{abstract}
In this work, we study the problem of learning a single model for multiple domains. Unlike the conventional machine learning scenario where each domain can have the corresponding model, multiple domains (i.e., applications/users) may share the same machine learning model due to maintenance loads in cloud computing services. For example, a digit-recognition model should be applicable to hand-written digits, house numbers, car plates, etc. Therefore, an ideal model for cloud computing has to perform well at each applicable domain. To address this new challenge from cloud computing, we develop a framework of robust optimization over multiple domains. In lieu of minimizing the empirical risk, we aim to learn a model optimized to the adversarial distribution over multiple domains. Hence, we propose to learn the model and the adversarial distribution simultaneously with the stochastic algorithm for efficiency. Theoretically, we analyze the convergence rate for convex and non-convex models. To our best knowledge, we first study the convergence rate of learning a robust non-convex model with a practical algorithm. Furthermore, we demonstrate that the robustness of the framework and the convergence rate can be further enhanced by appropriate regularizers over the adversarial distribution. The empirical study on real-world fine-grained visual categorization and digits recognition tasks verifies the effectiveness and efficiency of the proposed framework.
\end{abstract}

\section{Introduction}
\label{sec:intro}

Learning a single model for multiple domains becomes a fundamental problem in machine learning and has found applications in cloud computing services. Cloud computing witnessed the development of machine learning in recent years. Apparently, users of these cloud computing services can benefit from sophisticated models provided by service carrier, e.g., Aliyun. However, the robustness of deployed models becomes a challenge due to the explosive popularity of the cloud computing services. Specifically, to maintain the scalability of the cloud computing service, only a \textit{single} model will exist in the cloud for the same problem from different domains. For example, given a model for digits recognition in cloud, some users may call it to identify the handwritten digits while others may try to recognize the printed digits (e.g., house number). 

\begin{figure}[!ht]
\centering
\includegraphics[width=0.3\textwidth]{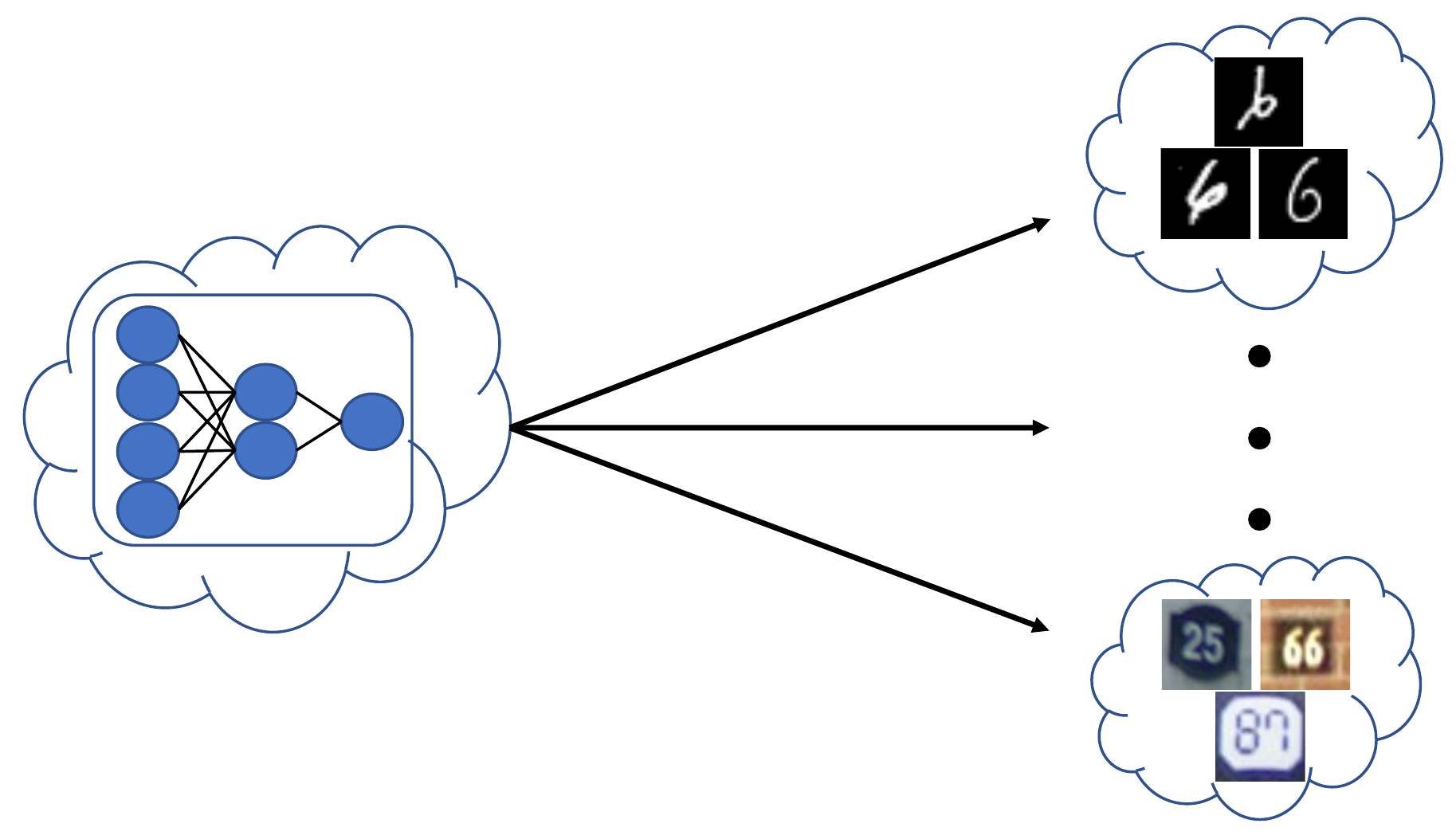}
\caption{Illustration of optimizing over multiple domains. In this example, a digit-recognition model provided by cloud service carrier should be applicable for multiple domains, e.g., handwritten digits, printed digits.}\label{fig:illus}
\end{figure} 

A satisfied model has to deal with both domains (i.e., handwritten digits, printed digits) well in the modern architecture of cloud computing services. This problem is illustrated in Fig.~\ref{fig:illus}. Note that the problem is different from multi-task learning~\cite{ZhangY17aa} that aims to learn different models (i.e., \textit{multiple} models) for different tasks by exploiting the shared information between related tasks.

In a conventional learning procedure, an algorithm may mix the data from multiple domains by assigning an ad-hoc weight for each example, and then learn a model accordingly. The weight is pre-defined and can be uniform for each example, which is known as empirical risk minimization (ERM). Explicitly, the learned model can handle certain domains well but perform arbitrarily poor on the others. The unsatisfied performance in certain domains will result in business interruption from users. Moreover, assigning even weights for all examples can suffer from the data imbalance problem when the examples from certain domains dominate.

Recently, distributionally robust optimization has attracted much attention~\cite{ChenLSS17,NamkoongD16,ShwartzW16}. Unlike the conventional strategy with the uniform distribution, it aims to optimize the performance of the model in the worst case distribution over examples. The learned model is explicitly more robust by focusing on the hard examples. To learn a robust model, many existing work apply the convex loss functions, while the state-of-the-art performance for several important practical problems are reported from the methods with non-convex loss functions, e.g, deep neural networks~\cite{HeZRS16,KrizhevskySH12,SzegedyLJSRAEVR15}. \cite{ChenLSS17} proposed an algorithm to solve the non-convex problem, but their analysis relies on a near-optimal oracle for the non-convex subproblem, which is not feasible for most non-convex problems in real tasks. Besides, their algorithm has to go through the whole data set at least once to update the parameters at every iteration, which makes it too expensive for the large-scale data set.

In this work, we propose a framework to learn a robust model over multiple domains rather than examples. By learning the model and the adversarial distribution simultaneously, the algorithm can balance the performance between different domains adaptively. Compared with the previous work, the empirical data distribution in each domain remains unchanged and our framework only learns the distribution over multiple domains. Therefore, the learned model will not be potentially misled by the adversarial distribution over examples. Our framework is also comparatively efficient due to the adoption of stochastic gradient descent (SGD) for optimization. More importantly, we first prove that the proposed method converges with a rate of $\OO(1/T^{1/3})$ without the dependency on the oracle. To further improve the robustness of the framework, we introduce a regularizer for the adversarial distribution. We find that an appropriate regularizer not only prevents the model from a trivial solution but also accelerates the convergence rate to $\OO(\sqrt{\log(T)/T})$. The detailed theoretical results are summarized in Table~\ref{ta:conv}. The empirical study on pets categorization and digits recognition demonstrates the effectiveness and efficiency of the proposed method.

\begin{table}[!ht]
\small
\centering
\caption{Convergence rate for the non-convex model and adversarial distribution (``Adv-Dist") under different settings.}\label{ta:conv}
\begin{tabular}{|l|l|l|l|}\hline
\multicolumn{2}{|l|}{Setting}&\multicolumn{2}{|l|}{Convergence}\\\hline
Model&Adv-Dist&Model&Adv-Dist\\\hline
Smooth&Concave&$\OO(\frac{1}{T^{1/3}})$&$\OO(\frac{1}{T^{1/3}})$\\\hline
Smooth&Strongly Concave&$\OO(\sqrt{\frac{log(T)}{T}})$&$\OO(\frac{log(T)}{T})$\\\hline
\end{tabular}
\end{table}


\section{Related Work}
\label{sec:related}

Robust optimization has been extensively studied in the past decades~\cite{BertsimasBC11}. Recently, it has been investigated to improve the performance of the model in the worst case data distribution, which can be interpreted as regularizing the variance~\cite{2016arXivDuchi}. For a set of convex loss functions (e.g., a single data set), \cite{NamkoongD16} and \cite{ShwartzW16} proposed to optimize the maximal loss, which is equivalent to minimizing the loss with the worst case distribution generated from the empirical distribution of data. \cite{NamkoongD16} showed that for the $f$-divergence constraint, a standard stochastic mirror descent algorithm can converge at the rate of $\OO(1/\sqrt{T})$ for the convex loss. In \cite{ShwartzW16}, the analysis indicates that minimizing the maximal loss can improve the generalization performance. In contrast to a single data set, we focus on dealing with multiple data sets and propose to learn the non-convex model in this work.

To tackle non-convex losses, \cite{ChenLSS17} proposed to apply a near-optimal oracle. At each iteration, the oracle is called to return a near-optimal model for the given distribution. After that, the adversarial distribution over examples is updated according to the model from the oracle. With an $\alpha$-optimal oracle, authors proved that the algorithm can converge to the $\alpha$-optimal solution at the rate of $\OO(1/\sqrt{T})$, where $T$ is the number of iterations. The limitation is that even if we assume a near-optimal oracle is accessible for the non-convex problem, the algorithm is too expensive for the real-world applications. It is because that the algorithm has to enumerate the whole data set to update the parameters at each iteration. Without a near-optimal oracle, we prove that the proposed method can converge with a rate of $\OO(\sqrt{\log(T)/T})$ with an appropriate regularizer and the computational cost is much cheaper.

\section{Robust Optimization over Multiple Domains}
\label{sec:method}
Given $K$ domains, we denote the data set as $\{S_1,\cdots,S_K\}$. For the $k$-th domain, $S_k = \{\x_i^k,y_i^k\}$, $\x_i^k$ is an example (e.g., an image) and $y_i^k$ is the corresponding label. We aim to learn a model that performs well over all domains. It can be cast as a robust optimization problem as follows.
\begin{eqnarray*}
&&\min_W \epsilon\\
s.t.&& \forall k, f_k(W)\leq \epsilon
\end{eqnarray*}
where $W$ is the parameter of a prediction model. $f_k(\cdot)$ is the empirical risk of the $k$-th domain as
\[f_k(W) = \sum_{i:\x_i^k\in S_k}\frac{1}{|S_k|}\ell(\x_i^k, y_i^k;W)\]
and $\ell(\cdot)$ can be any non-negative loss function. Since the cross entropy loss is popular in deep learning, we will adopt it in the experiments.

The problem is equivalent to the following minimax problem
\begin{eqnarray}\label{eq:problem}
\min_W\max_{\p:\p\in \Delta} \LL(\p,W)= \p^\top \f(W)
\end{eqnarray}
where $\f(W) = [f_1(W),\cdots,f_K(W)]^\top$. $\p$ is an adversarial distribution over multiple domains and $\p\in \Delta$, where $\Delta$ is the simplex as $\Delta = \{\p\in\R^{K}|\sum_{k=1}^K p_k=1; \forall k,\ p_k\geq 0\}$.

It is a game between the prediction model and the adversarial distribution. The minimax problem can be solved in an alternating manner, which applies gradient descent to learn the model and gradient ascent to update the adversarial distribution. Considering the large number of examples in each data set, we adopt SGD to observe an unbiased estimation for the gradient at each iteration, which avoids enumerating the whole data set. Specifically, at the $t$-th iteration, a mini-batch of size $m$ is randomly sampled from each domain. The loss of the mini-batch from the $k$-th domain is
\[\hat{f}_k^t(W) = \frac{1}{m}\sum_{i=1}^m \ell(\hat{\x}_{i:t}^k,\hat{y}_{i:t}^k;W)\]
It is apparent that $E[\hat{f}^t_k(W)] = f_k(W)$ and $E[\nabla \hat{f}^t_k(W)] = \nabla f_k(W)$.
 
 \begin{algorithm}[!h]
   \caption{Stochastic Algorithm for Robust Optimization}
   \label{alg:conv}
\begin{algorithmic}
   \STATE {\bfseries Input:} Data set $\{S_1,\cdots, S_K\}$, size of mini-batch $m$, step-sizes $\eta_w$, $\eta_p$
   \STATE Initialize $\p_1 = [1/K,\cdots,1/K]$
   \FOR{$t=1$ {\bfseries to} $T$}
   \STATE Randomly sample $m$ examples from each domain
   \STATE Update $W_{t+1}$ as in Eqn.~\ref{eq:updatew}
   \STATE Update $\p_{t+1}$ as in Eqn.~\ref{eq:updatep}
   \ENDFOR
   \RETURN $\overline{W} = \frac{1}{T}\sum_t W_t$, $\bar{\p} = \frac{1}{T}\sum_t \p_t$
\end{algorithmic}
\end{algorithm}

After sampling, we first update the model by gradient descent as
\begin{align}\label{eq:updatew}
&W_{t+1} = W_t - \eta_w \hat{g}_t; \ \ \text{where}\ \  \hat{g}_t = \sum_k p_k^t\nabla \hat{f}_k^t(W_t)
\end{align}

Then, the distribution $\p$ is updated in an adversarial way. Since $\p$ is from the simplex, we can adopt multiplicative updating criterion~\cite{AroraHK12} to update it as
\begin{eqnarray}\label{eq:updatep}
&&p_{t+1}^k = \frac{p_t^k\exp(\eta_p \hat{f}_k^t(W_{t}))}{Z_t};\nonumber\\
&& \text{where}\quad Z_t = \sum_k p_t^k\exp(\eta_p \hat{f}_k^t(W_{t}))
\end{eqnarray}

Alg.~\ref{alg:conv} summarizes the main steps of the approach. For the convex loss functions, the convergence rate is well known~\cite{NemirovskiJLS09} and we provide a high probability bound for completeness. All detailed proofs of this work can be found in the appendix.
\begin{lemma}\label{thm:convex}
Assume the gradient of $W$ and the function value are bounded as $\forall t$, $\|\nabla\hat{f}_k^t(W_t)\|_F\leq \sigma$, $\|\hat{\f}^t(W_t)\|_2\leq \gamma$ and $\forall W,\ \|W\|_F\leq R$. Let $(\overline{W},\bar{\p})$ denote the results returned by Alg.~\ref{alg:conv} after $T$ iterations. Set the step-sizes as 
$\eta_w = \frac{R}{\sigma\sqrt{T}}$ and $\eta_p = \frac{2\sqrt{2\log(K)}}{\gamma\sqrt{T}}$. Then, with a probability $1-\delta$, we have
\[\max_{\p}\LL(\p,\overline{W}) - \min_W\LL(\bar{\p},W)\leq \frac{c_1}{\sqrt{T}}+\frac{2c_2\sqrt{\log(2/\delta)}}{\sqrt{T}}\]
where $c_1=\OO(\sqrt{\log(K)})$ and $c_2$ is a constant.
\end{lemma}
Lemma~\ref{thm:convex} shows that the proposed method with the convex loss can converge to the saddle point at the rate of $\OO(1/\sqrt{T})$ with high probability, which is a stronger result than the expectation bound in \cite{NamkoongD16}. Note that setting $\eta_w = \OO(\frac{1}{\sqrt{T}})$ and $\eta_{p} = \OO(\sqrt{\frac{\log(K)}{T}})$ will not change the order of the convergence rate, which means $\sigma$, $\gamma$ and $R$ are not required for implementation.

\subsection{Non-convexity}
Despite the extensive studies about the convex loss, there is little research about the minimax problem with non-convex loss. To provide the convergence rate for the non-convex problem, we first have the following lemma.
\begin{lemma}\label{lem:nonconvex}
With the same assumptions as in Lemma~\ref{thm:convex}, if $\ell(\cdot)$ is non-convex but $L$-smoothness, we have
\footnotesize{
\begin{align*}
&\sum_tE[\|\nabla_{W_t} \LL(\p_t, W_t)\|_F^2]\leq \frac{\LL(\p_0,W_0)}{\eta_w}+\frac{\eta_pT\gamma^2}{2\eta_w}+\frac{TL\eta_w \sigma^2}{2}\\
& \sum_t E[\LL(\p_t,W_t)]\geq \max_{\p\in\Delta}\sum_t E[\LL(\p,W_t)] -( \frac{\log(K)}{\eta_p}+\frac{T\eta_p\gamma^2}{8})
\end{align*}}
\end{lemma}

Since the loss is non-convex, the convergence is measured by the norm of the gradient (i.e., stationary point), which is a standard criterion for the analysis in the non-convex problem~\cite{GhadimiL13a}. Lemma~\ref{lem:nonconvex} indicates that $W$ can converge to a stationary point where $\p_t$ is a qualified adversary by setting the step-sizes elaborately. Furthermore, it demonstrates that the convergence rate of $W$ will be influenced by the convergence rate of $\p$ via $\eta_p$.

With Lemma~\ref{lem:nonconvex}, we have the convergence analysis of the non-convex minimax problem as follows.
\begin{thm}\label{thm:nonconvex}
With the same assumptions as in Lemma~\ref{lem:nonconvex}, if we set the step-sizes as
$\eta_w=\frac{\sqrt{2\gamma\sqrt{2\log(K)}}}{\sigma\sqrt{L}}T^{-1/3}$ and $\eta_p=\frac{2\sqrt{2\log(K)}}{\gamma}T^{-2/3}$, we have
\footnotesize{
\begin{align*}
&E[\frac{1}{T}\sum_t\| \nabla_{W_t}\LL(\p_t,W_t)\|_F^2]\\
&\leq (\frac{\LL(\p_0,W_0)}{\sqrt{2\gamma \sqrt{2\log(K)}}}+\sqrt{2\gamma \sqrt{2\log(K)}})\sigma\sqrt{L} T^{-1/3}\\
&E[\frac{1}{T}\sum_t \LL(\p_t,W_t)]\\
&\geq E[\max_{\p\in\Delta}\frac{1}{T}\sum_t\LL(\p,W_t)]- \frac{\gamma\sqrt{\log(K)}}{\sqrt{2}}T^{-1/3}
\end{align*}}
\end{thm}
\paragraph{Remark} Compared with the convex case in Lemma~\ref{thm:convex}, the convergence rate of a non-convex problem is degraded from $\OO(1/\sqrt{T})$ to $\OO(1/T^{1/3})$. It is well known that the convergence rate of general minimization problems with a smooth non-convex loss can be up to $\OO(1/\sqrt{T})$~\cite{GhadimiL13a}. Our results further demonstrate that minimax problems with non-convex loss is usually harder than non-convex minimization problems.

Different step-sizes can lead to different convergence rates. For example, if the step-size for updating $\p$ is increased as $\eta_p = 1/\sqrt{T}$ and that for model is decreased as $\eta_w=1/T^{1/4}$, the convergence rate of $\p$ can be accelerated to $\OO(1/\sqrt{T})$ while the convergence rate of $W$ will degenerate to $\OO(1/T^{1/4})$. Therefore, if a sufficiently small step-size is applicable for $\p$, the convergence rate of $W$ can be significantly improved. We exploit this observation to enhance the convergence rate in the next subsection.

\subsection{Regularized Non-convex Optimization}
\label{sec:robust}
A critical problem in minimax optimization is that the formulation is very sensitive to the outlier. For example, if there is a domain with significantly worse performance than others, it will dominate the learning procedure according to Eqn.~\ref{eq:problem} (i.e., one-hot value in $\p$). Besides the issue of robustness, it is prevalent in real-world applications that the importance of domains is different according to their budgets, popularity, etc. Incorporating the side information into the formulation is essential for the success in practice. Given a prior distribution, the problem can be written as
\begin{eqnarray*}
&&\min_W\max_{\p:\p\in \Delta}\p^\top \f(W)\\
s.t.&& \D(\p||\q)\leq \tau
\end{eqnarray*}
where $\q$ is the prior distribution which can be a distribution defined from the side information or a uniform distribution for robustness. $\D(\cdot)$ defines the distance between two distributions, e.g., $L_p$ distance or $\mathrm{KL}$-divergence
\begin{align*}
&\D_{L_2}(\p||\q) = \|\p-\q\|_2^2;\quad \D_{\mathrm{KL}}(\p||\q) = \sum_k p_k\log(p_k/q_k)
\end{align*}
Since $\mathrm{KL}$-divergence cannot handle the prior distribution with zero elements, optimal transportation (OT) distance becomes popular recently to overcome the drawback
\[\D_{\mathrm{OT}}(\p||\q) = \min_{P\in U(\p,\q)}\langle P,M\rangle\]
For computational efficiency, we use the version with an entropy regularizer~\cite{Cuturi13} and we have
\begin{prop}\label{prop:ot}
Define the $\mathrm{OT}$ regularizer as 
\begin{eqnarray}\label{eq:otr}
&&\D_{\mathrm{OT}}(\p||\q) =\max_{\alpha,\beta} \min_P\frac{1}{\nu} \sum_{i,j} P_{i,j}\log(P(i,j))\nonumber\\
&&+ P_{i,j}M_{i,j}+\alpha^\top(P\mathbf{1}_K - \p)+\beta^\top(P\mathbf{1}_K-\q)
\end{eqnarray}
and it is convex in $\p$.
\end{prop}

According to the duality theory~\cite{boyd2004convex}, for each $\tau$, we can have the equivalent problem with a specified $\lambda$
\begin{eqnarray}\label{eq:problemr}
\min_W\max_{\p:\p\in \Delta} \hat{\LL}(\p,W)= \p^\top \f(W)-\frac{\lambda}{2} \D(\p||\q)
\end{eqnarray}
Compared with the formulation in Eqn.~\ref{eq:problem}, we introduce a regularizer for the adversarial distribution.

If $\D(\p||\q)$ is convex in $\p$, the similar convergence as in Theorem.~\ref{thm:nonconvex} can be obtained with the same analysis. Moreover, according to the research for SGD, the strongly convexity is the key to  achieve the optimal convergence rate~\cite{RakhlinSS12}. Hence, we adopt a strongly convex regularizer i.e., $L_2$ regularizer, for the distribution. The convergence rate for other strongly convex regularizers can be obtained with a similar analysis by defining the smoothness and the strongly convexity with the corresponding norm.

Equipped with the $L_2$ regularizer, the problem in Eqn.~\ref{eq:problemr} can be solved with projected first-order algorithm. We adopt the projected gradient ascent to update the adversarial distribution as
\begin{eqnarray*}
\p_{t+1} = \P_{\Delta}(\p_t + \eta_p^t \hat{h}^t);\quad \text{where}\quad \hat{h}^t =  \hat{f}^t - \lambda (\p_t - \q)
\end{eqnarray*}
$\P_{\Delta}(\p)$ projects the vector $\p$ onto the simplex. The projection algorithm can be found in \cite{DuchiSSC08} which is based on $K.K.T.$ condition. We also provide the gradient of $\mathrm{OT}$ regularizer in the appendix.

Since the regularizer (i.e., $-L_2$) is strongly concave, the convergence of $\p$ can be accelerated dramatically, which leads to a better convergence rate for the minimax problem. The theoretical result is as follows.
\begin{thm}\label{thm:regularizer}
With the same assumptions as in Theorem~\ref{thm:nonconvex}, if we assume $\forall t,\ \|\hat{h}^t\|_2\leq \mu$ and set step-sizes as
$\eta_w = \frac{2\mu\sqrt{\log(T)}}{\sigma\sqrt{\lambda L T}}$ and $\eta_p^t=\frac{1}{\lambda t}$, we have
\footnotesize{
\begin{align*}
&E[\frac{1}{T}\sum_t\| \nabla_{W_t}\hat{\LL}(\p_t,W_t)\|_F^2]\\
&\leq \left(\frac{\LL(\p_0,W_0)\sigma \sqrt{\lambda L}}{2\mu\sqrt{\log(T)}}+\frac{\mu\pi^2\sigma\sqrt{\lambda L}}{12}+2\mu\sigma\sqrt{\lambda L\log(T)}\right)\frac{1}{\sqrt{T}}\\
&E[\frac{1}{T}\sum_t \hat{\LL}(\p_t,W_t)]\geq E[\max_{\p\in\Delta}\frac{1}{T}\sum_t\hat{\LL}(\p,W_t)]- \frac{\mu^2\log(T)}{\lambda T}
\end{align*}}
\end{thm}
\paragraph{Remark} With the strongly concave regularizer, it is not surprise to obtain the $\OO(\log(T)/T)$ convergence rate for $\p$. As we discussed in Lemma~\ref{lem:nonconvex}, a fast convergence rate of $\p$ can improve that of $W$. In Theorem~\ref{thm:regularizer}, the convergence rate of $W$ is improved from $\OO(1/T^{1/3})$ to $\OO(\sqrt{\log(T)/T})$. It shows that the applied regularizer not only improves the robustness of the proposed framework but also accelerates the learning procedure.

Moreover, the step-size for the adversarial distribution provides a trade-off between the bias and variance of the gradient. Therefore, the convergence rate can be further improved by reducing the variance. We shrink the gradient with a factor $c$ and update the distribution as
\begin{eqnarray*}
\p_{t+1} = \P_{\Delta}(\p_t + \frac{\eta_p^t}{1+c/t} \hat{h}^t)
\end{eqnarray*}
When taking $\eta_p^t=\frac{1}{\lambda t}$, the update becomes 
\begin{eqnarray}\label{update:pr}
\p_{t+1} = \P_{\Delta}(\p_t+\frac{1}{\lambda(t+c)}\hat{h}^t)
\end{eqnarray}
With a similar analysis as Theorem~\ref{thm:regularizer}, we have
\begin{thm}\label{thm:regularizer2}
With the same assumptions as in Theorem~\ref{thm:regularizer}, if we set the step-size $\eta_p^t = \frac{1}{\lambda(t+c)}$, we have
\begin{align*}
&E[\frac{1}{T}\sum_t \hat{\LL}(\p_t,W_t)]\\
&\geq E[\max_{\p\in\Delta}\frac{1}{T}\sum_t\hat{\LL}(\p,W_t)]- (\lambda c+ \frac{\mu^2}{2\lambda}\ln(\frac{T}{c}+1)+\frac{\mu^2}{2\lambda})\frac{1}{T}
\end{align*}
\end{thm} 
It shows that the constant $c$ can control the trade-off between bias (i.e., $\lambda c$) and variance (i.e., $ \frac{\mu^2}{2\lambda}\ln(\frac{T}{c}+1)$). By setting the constant appropriately, we can have the following corollary
\begin{cor}\label{cor:constant}
When setting $c = \frac{\mu^2}{\lambda^2(1+\sqrt{1+\frac{2\mu^2}{\lambda^2T}})}$, the RHS in Theorem~\ref{thm:regularizer2} is maximum.
\end{cor}
The optimality is from the fact that RHS is concave in $c$ and detailed discussion can be found in the appendix. 


The algorithm for robust optimization with the regularizer is summarized in Alg.~\ref{alg:nonconv}.
\begin{algorithm}[!h]
   \caption{Stochastic Regularized Robust Optimization}
   \label{alg:nonconv}
\begin{algorithmic}
   \STATE {\bfseries Input:} Data set $\{S_1,\cdots, S_K\}$, size of mini-batch $m$, step-sizes $\eta_w$, $\eta_p$
   \STATE Initialize $\p_1 = [1/K,\cdots,1/K]$
   \STATE Compute the constant $c$ as in Corollary~\ref{cor:constant}
   \FOR{$t=1$ {\bfseries to} $T$}
   \STATE Randomly sample $m$ examples from each domain
   \STATE Update $W_{t+1}$ with gradient descnet
   \STATE (Optional) Solve the problem in Eqn.~\ref{eq:otr} if applying $\D_{\mathrm{OT}}(\p_t||\q)$
   \STATE Update $\p_{t+1}$ with gradient ascent
   \STATE Project $\p_{t+1}$ onto the simplex
   \ENDFOR
\end{algorithmic}
\end{algorithm}

\subsection{Trade Efficiency for Convergence}
In this subsection, we study if we can recover the optimal convergence rate for the general non-convex problem as in \cite{GhadimiL13a}. Note that \cite{ChenLSS17} applies a near-optimal oracle to achieve the $\OO(1/\sqrt{T})$ convergence rate. Given a distribution, it is hard to observe an oracle for the non-convex model. In contrast, obtaining the near-optimal adversarial distribution with a fixed model is feasible. For the original problem in Eqn.~\ref{eq:problem}, the solution is trivial as returning the index of the domain with the largest empirical loss. For the problem with the regularizer in Eqn.~\ref{eq:problemr}, the near-optimal $\p$ can be obtained efficiently by any first order methods~\cite{boyd2004convex}. Therefore, we can change the updating criterion for the distribution at the $t$-th iteration to
\begin{eqnarray}\label{eq:costup}
&&\text{Obtain }\p_{t+1}\text{ such that }\|\p_{t+1}-\p_{t+1}^*\|_1\leq \xi_{t+1} \nonumber\\
&& \text{where}\quad \p_{t+1}^*= \arg\max_{\p:\p\in\Delta} \LL(\p,W_t)
\end{eqnarray}
With the new updating criterion and letting $\F(W) = \max_{\p}\LL(\p, W)$, we can have a better convergence rate as follows.
\begin{thm}\label{thm:opt}
With the same assumptions as in Theorem~\ref{thm:nonconvex}, if we update $\p$ as in Eqn.~\ref{eq:costup}, where $\xi_t = \frac{1}{\sqrt{t}}$, and set the step-size as $\eta_w = \frac{\sqrt{2}}{\sigma\sqrt{L T}}$, we have
\begin{align*}
\sum_t E[\frac{1}{T}\|\nabla \F(W_t)\|_F^2] \leq (\F(W_0)+1)\frac{\sqrt{L}\sigma}{\sqrt{2T}}+\frac{2\sigma^2}{\sqrt{T}}
\end{align*}
\end{thm}
For the problem in Eqn.~\ref{eq:problem}, $\xi_t$ can be $0$ by a single pass through the whole data set. It shows that with an expensive but feasible operator as in Eqn.~\ref{eq:costup}, the proposed method can recover the optimal convergence rate for the non-convex problem.

\section{Experiments}
\label{sec:exp}
We conduct the experiments on training deep neural networks over multiple domains. The methods in the comparison are summarized as follows.
\begin{itemize}
\item \textbf{Individual}: It learns the model from an individual domain.
\item \textbf{Mixture$_{\mathrm{Even}}$}: It learns the model from multiple domains with even weights, which is equivalent to fixing $\p$ as an uniform distribution.
\item \textbf{Mixture$_{\mathrm{Opt}}$}: It implements the approach proposed in Alg.~\ref{alg:nonconv} that learns the model and the adversarial distribution over multiple domains simultaneously.
\end{itemize}
We adopt the popular cross entropy loss as the loss function $\ell(\cdot)$ in this work. Deep models are trained with SGD and the size of each mini-batch is set to $200$. For the methods learning with multiple domains, the number of examples from different domains are the same in a mini-batch and the size is $m=200/K$. Compared with the strategy that samples examples according to the learned distribution, the applied strategy is deterministic and will not introduce extra noise. The method is evaluated by investigating the worst case performance among multiple domains. For the worst case accuracy, it is defined as $\mathrm{Acc}_w = \min_k\{\mathrm{Acc}_1,\cdots,\mathrm{Acc}_K\}$. The worst case loss is defined as $f_w(W) = \max_k\{f_1(W),\cdots,f_K(W)\}$. All experiments are implemented on an NVIDIA Tesla P100 GPU.

\begin{figure*}[!ht]
\centering
\begin{minipage}{1.7in}
\centering
\includegraphics[height= 1.2in, width=1.6in ]{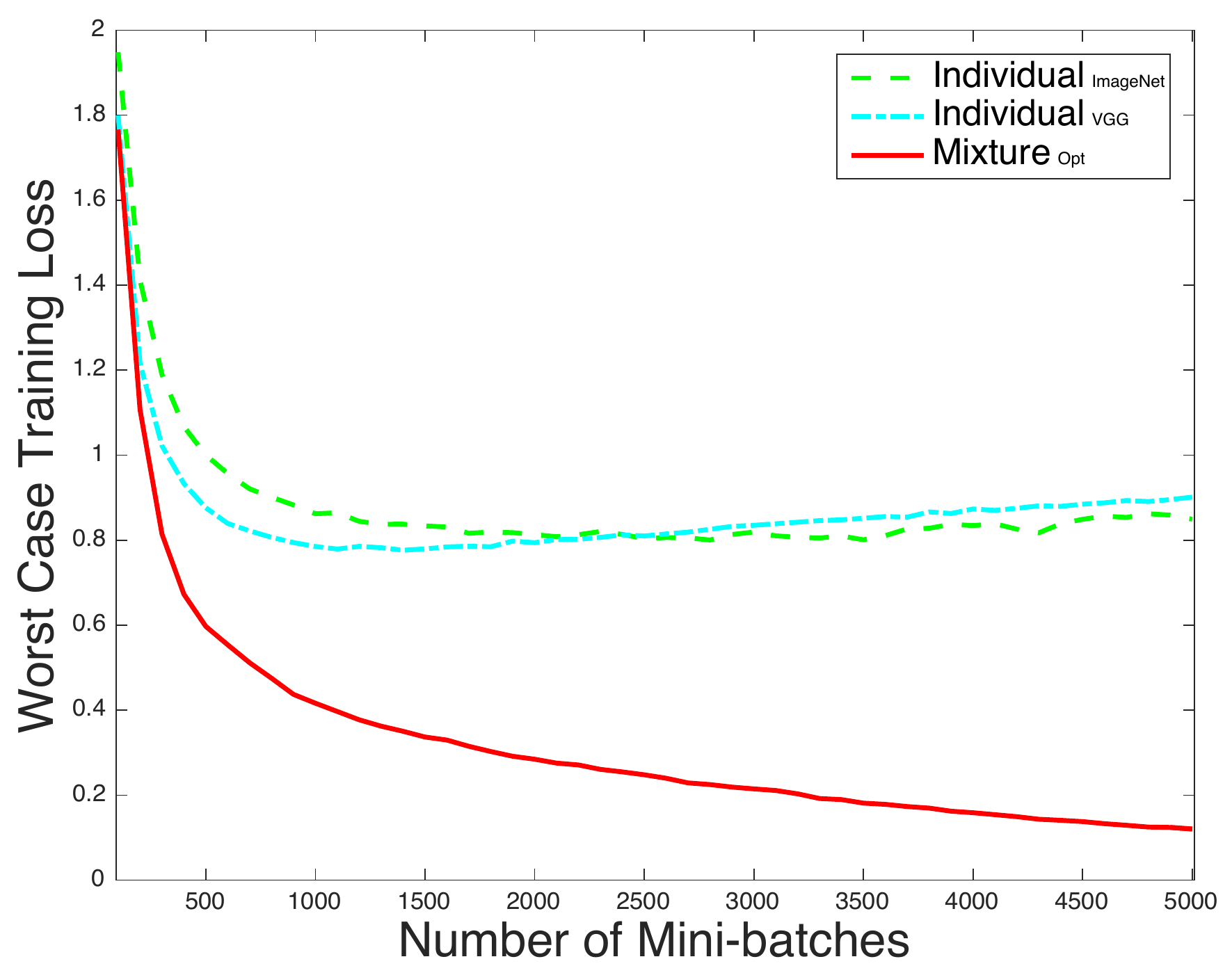}
\mbox{\footnotesize (a) Pets Categorization  }
\end{minipage}
\begin{minipage}{1.7in}
\centering
\includegraphics[height= 1.2in, width=1.6in ]{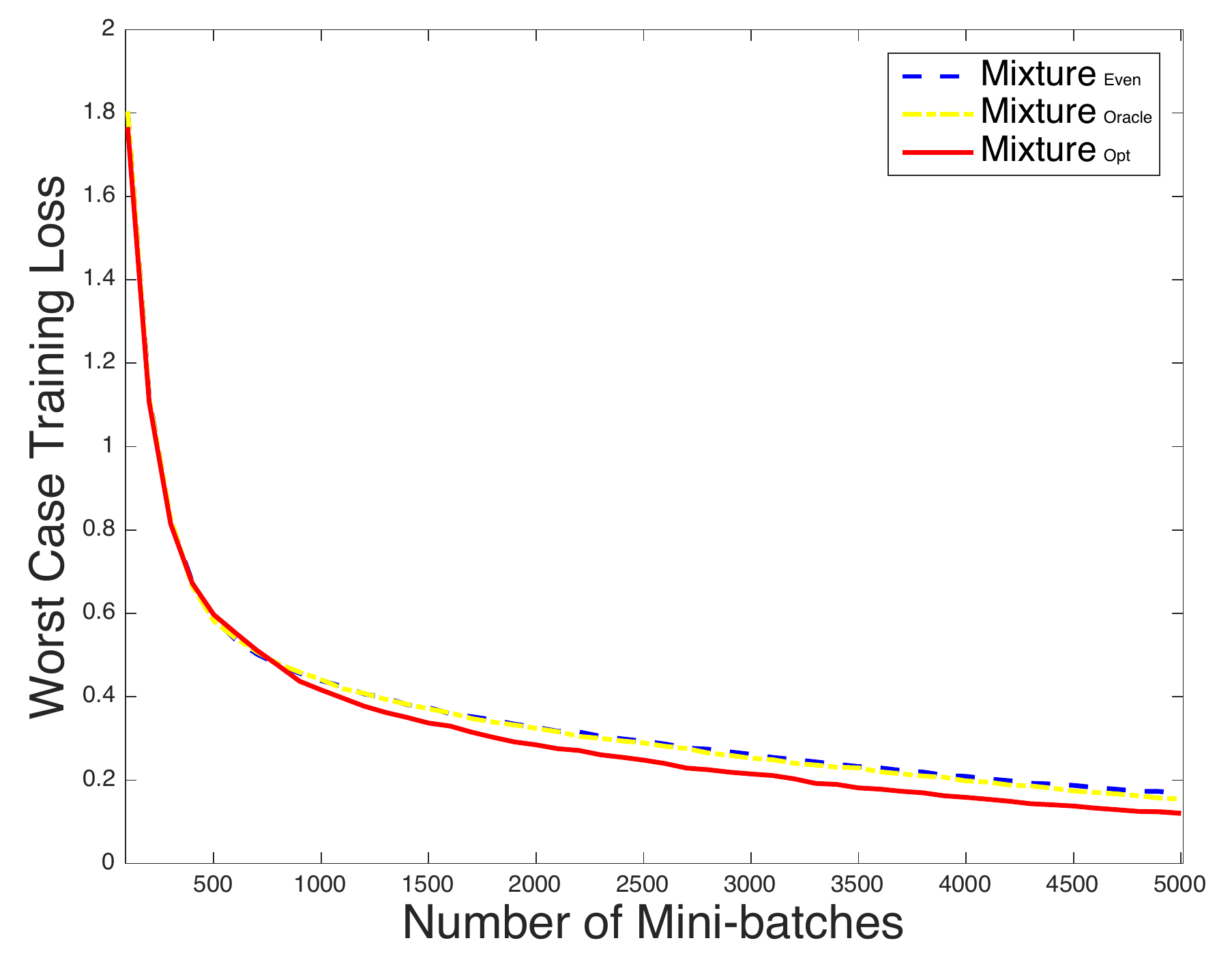}
\mbox{\footnotesize (b) Pets Categorization}
\end{minipage}
\begin{minipage}{1.7in}
\centering
\includegraphics[height= 1.2in, width=1.6in ]{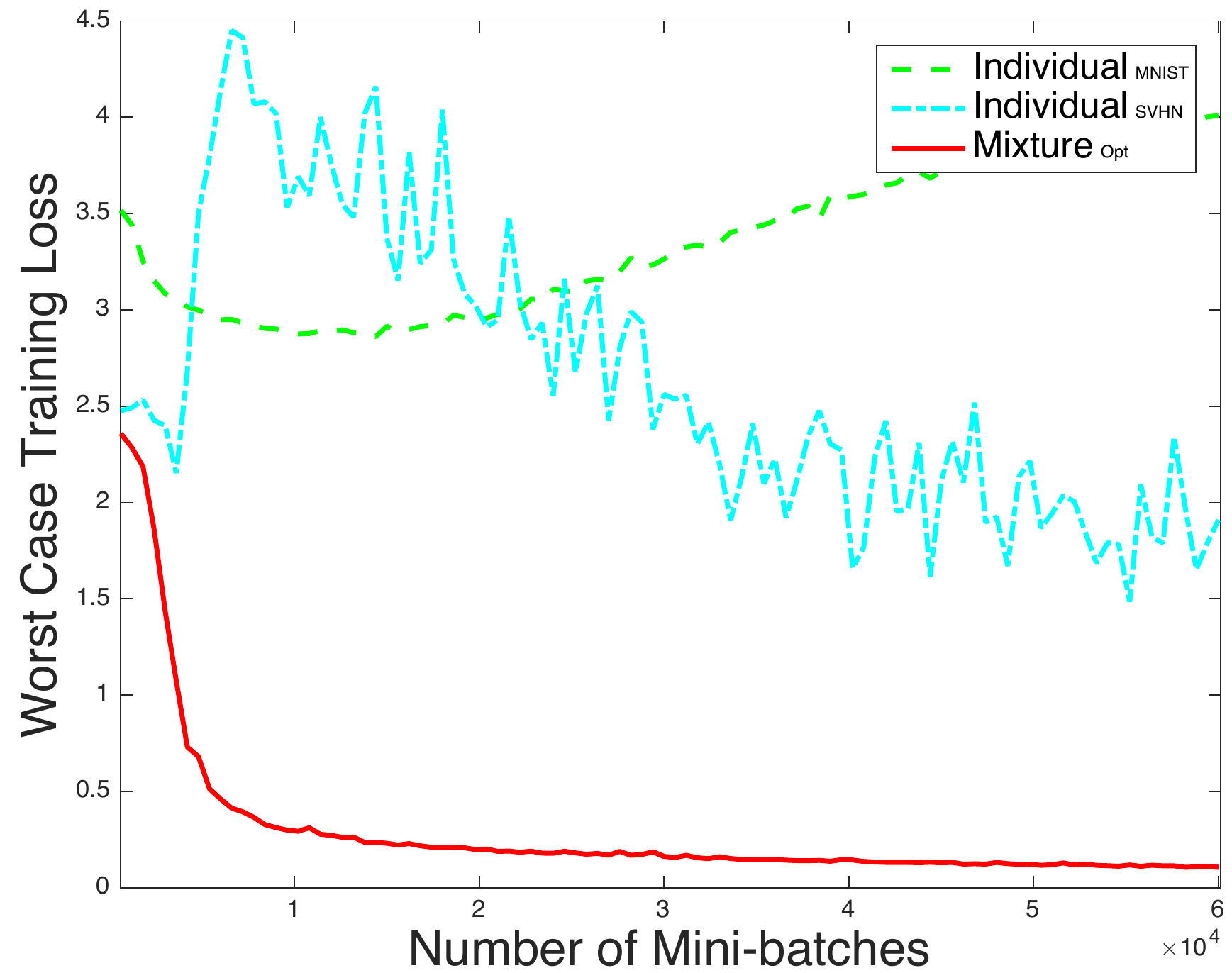}
\mbox{\footnotesize (c) Digits Recognition }
\end{minipage}
\begin{minipage}{1.7in}
\centering
\includegraphics[height= 1.2in, width=1.6in ]{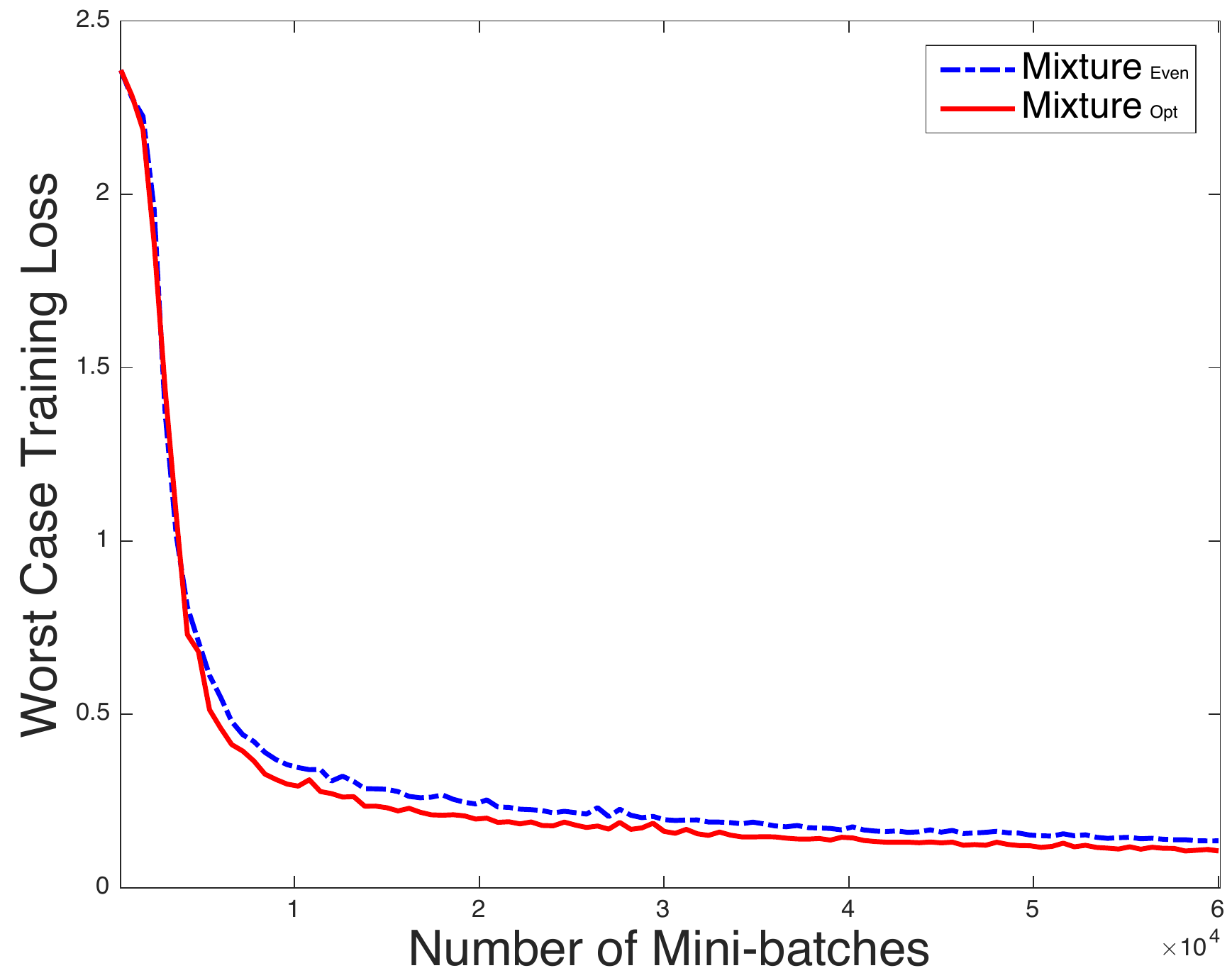}
\mbox{\footnotesize (d) Digits Recognition }
\end{minipage}
\caption{\label{fig:tloss}Illustration of worst case training loss.}
\end{figure*}

\begin{table*}[!ht]
\small
\centering
\caption{Comparison on pets categorization. We report the loss and accuracy ($\%$) on each data set.}\label{ta:fgvc}
\begin{tabular}{|l||l|l|l|l|l|l|l|l|l|l|l|l|}\hline
\multirow{2}{*}{Methods} &
\multicolumn{3}{l|}{ImageNet} &\multicolumn{3}{l|}{VGG}&\multirow{2}{*}{Acc$_{\mathrm{Tr}_\mathrm{w}}$}&\multirow{2}{*}{Acc$_{\mathrm{Te}_\mathrm{w}}$}\\
\cline{2-7}&Loss$_{\mathrm{Tr}}$&Acc$_{\mathrm{Tr}}$ &Acc$_{\mathrm{Te}} $&Loss$_{\mathrm{Tr}}$&Acc$_{\mathrm{Tr}}  $&Acc$_{\mathrm{Te}} $&&\\\hline
Individual$_{\mathrm{ImageNet}}$&$0.07$&$98.95$&$89.92$&$0.85$&$74.56$&$80.44$&$74.56$&$80.44$\\\hline
Individual$_{\mathrm{VGG}}$&$0.90$&$75.47$&$77.92$&$0.02$&$100.00$&$86.85$&$75.47$&$77.92$\\\hline
Mixture$_{\mathrm{Even}}$&$0.17$&$95.56$&$88.50$&$0.05$&$99.58$&$89.85$&$95.56$&$88.50$\\\hline\hline
Mixture$_{\mathrm{Oracle}}$&$0.15$&$96.04$&$88.92$&$0.06$&$99.41$&$89.99$&$96.04$&$88.92$\\\hline
Mixture$_{\mathrm{Opt}}$&$0.12$&$97.36$&$89.42$&$0.11$&$97.72$&$89.35$&$\textbf{97.36}$&$\textbf{89.35}$\\\hline
\end{tabular}
\end{table*}

\subsection{Pets Categorization}
\begin{figure}[!ht]
\centering
\begin{minipage}{3.4in} 
\centering
\includegraphics[height= 1.2 in, width=2.8in ]{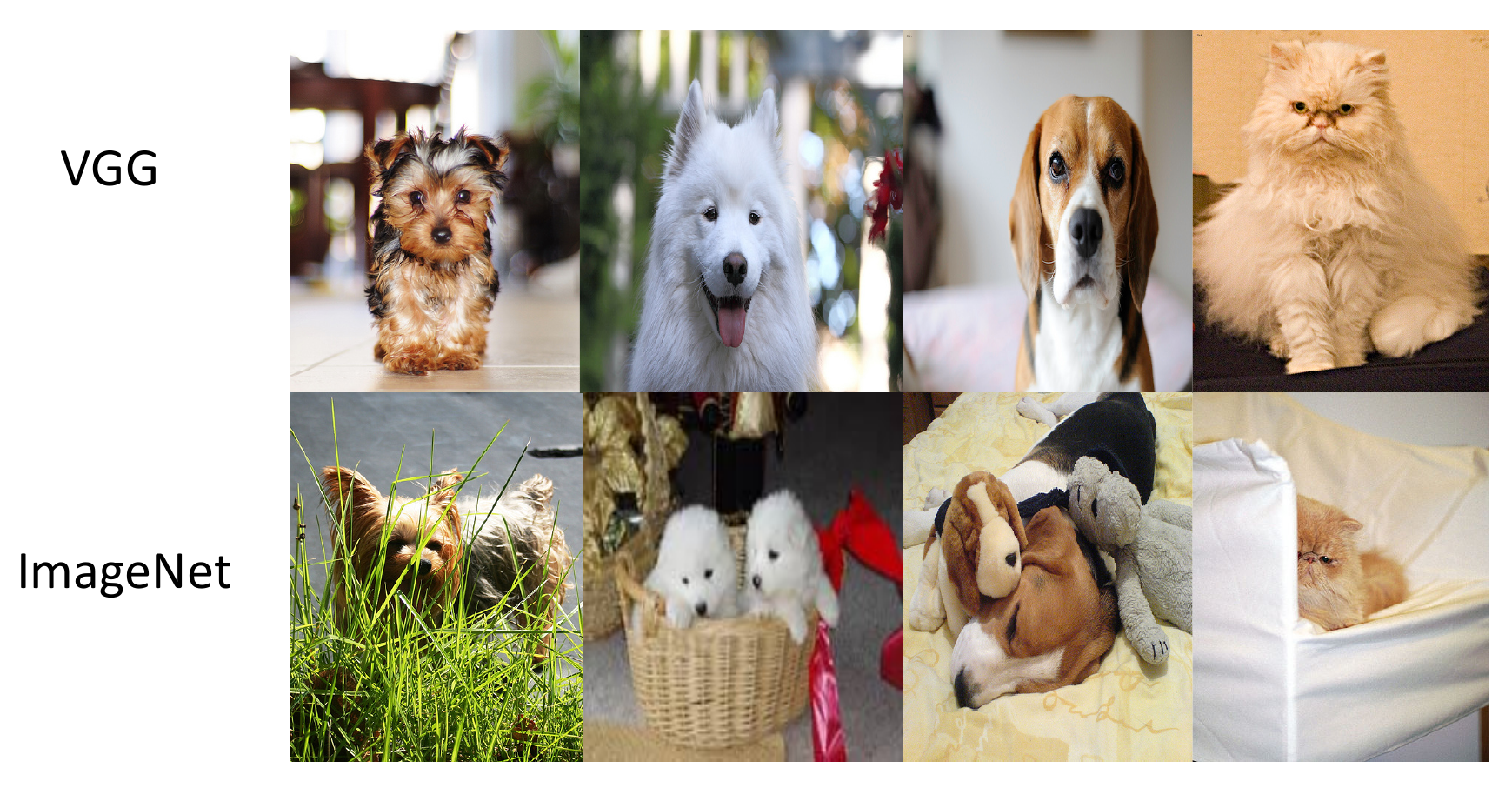}
\captionsetup{font={small}}
\caption{\small{Exemplar images from ImageNet and VGG.}}\label{fig:data}
\end{minipage}

\begin{minipage}{1.6in} 
\centering
\includegraphics[height= 1.1in, width=1.5in ]{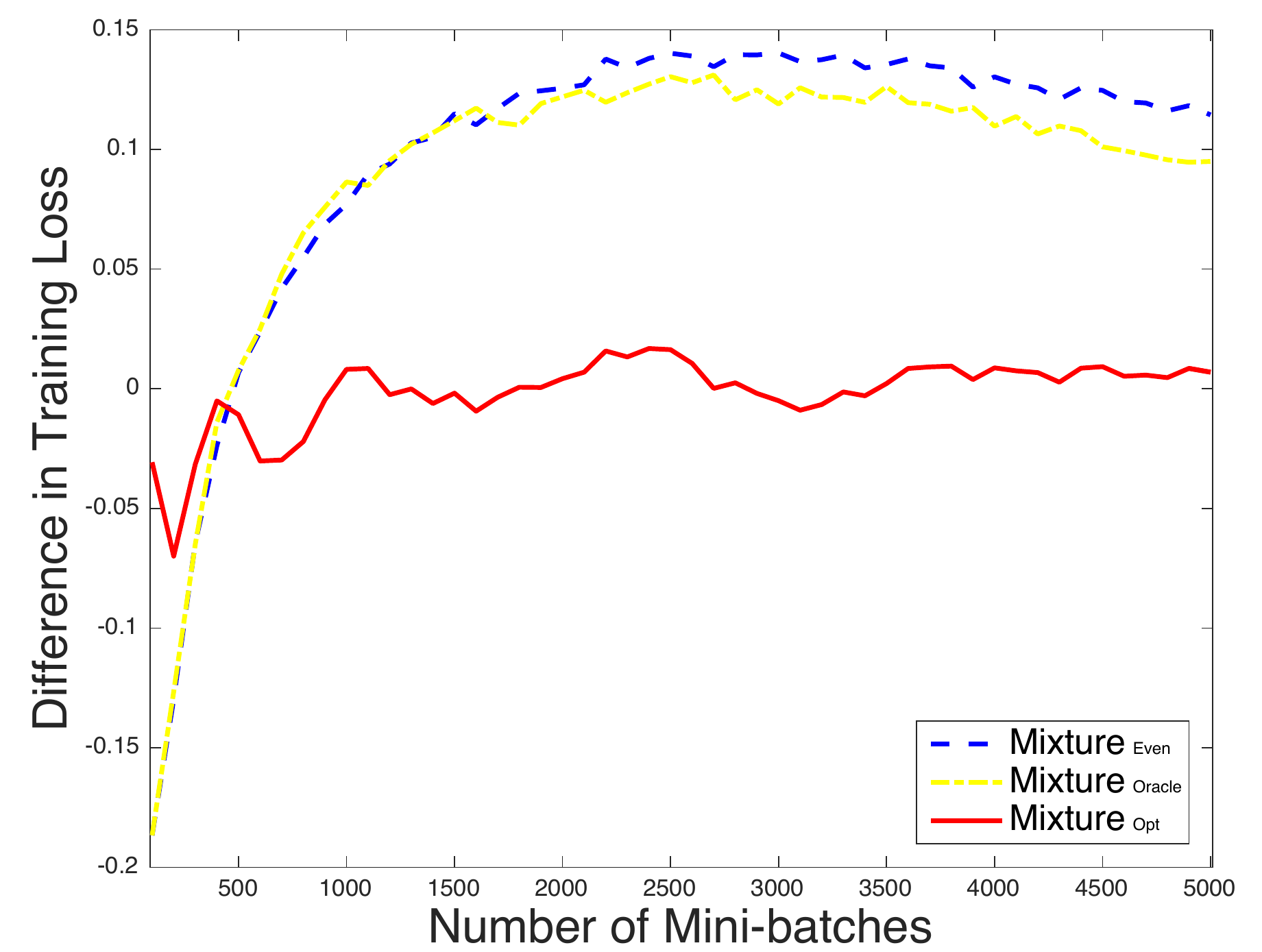}
\captionsetup{font={small}}
\caption{\small{Comparison of discrepancy in losses.}}\label{fig:diffcd}
\end{minipage}
\begin{minipage}{1.6in} 
\centering
\includegraphics[height= 1.1in, width=1.5in ]{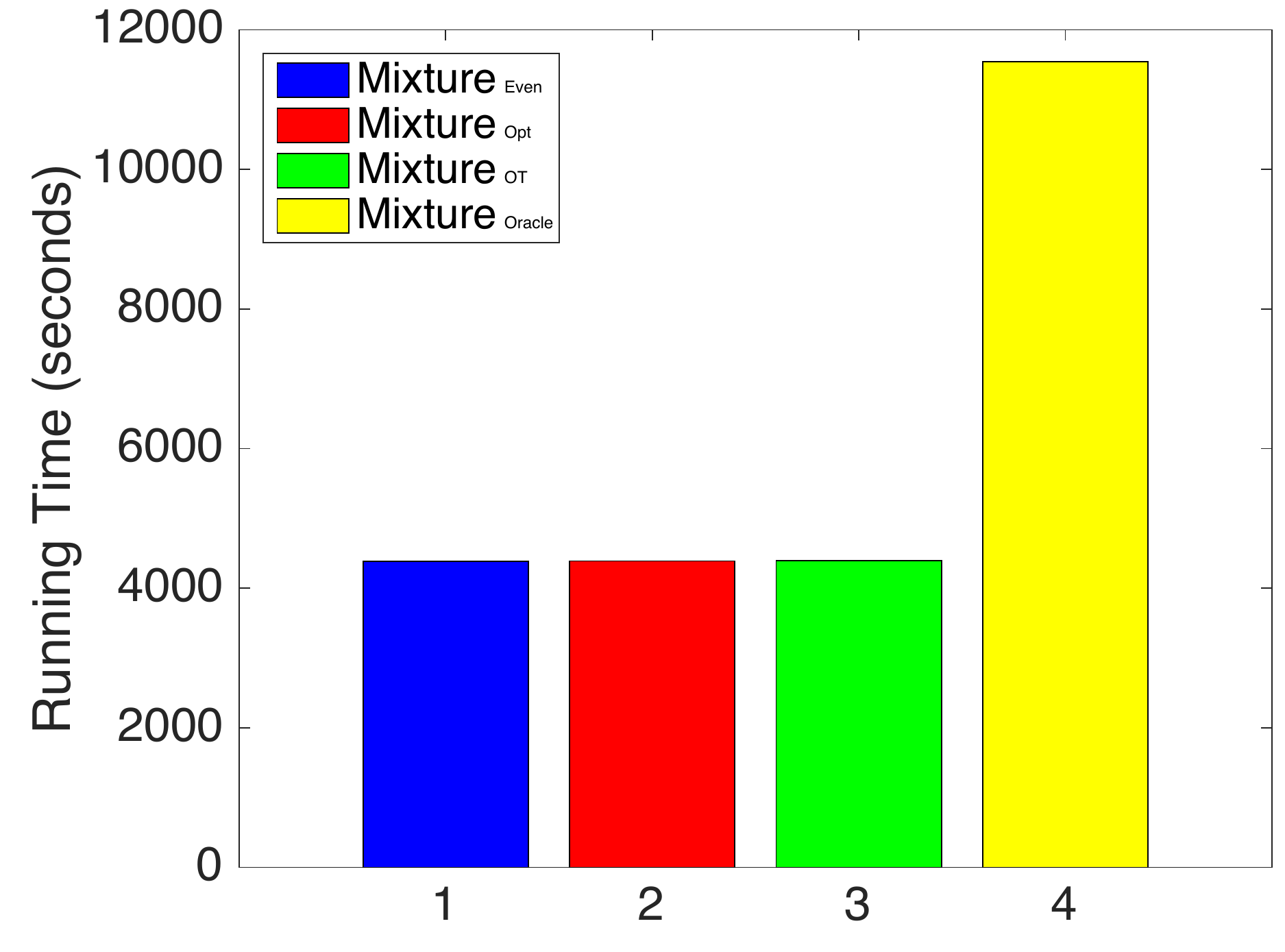}
\captionsetup{font={small}}
\caption{\small{Comparison of running time.}}\label{fig:rtime}
\end{minipage}
\end{figure}

First, we compare the methods on a fine-grained visual categorization task. Given the data sets of VGG cats\&dogs~\cite{parkhi12a} and ImageNet~\cite{ILSVRC15}, we extract the shared labels between them and then generate the subsets with desired labels from them, respectively. The resulting data set consists of 24 classes and the task is to assign the image of pets to one of these classes. For ImageNet, each class contains about $1,200$ images for training while that of VGG only has $100$ images. Therefore, we apply data augmentation by flipping (horizontal+vertical) and rotating ($\{45^\circ,\cdots, 315^\circ\}$) for VGG to avoid overfitting. After that, the number of images in VGG is similar to that of ImageNet. Some exemplar images from these data sets are illustrated in Fig.~\ref{fig:data}. We can find that the task in ImageNet is more challenging than that in VGG due to complex backgrounds.

\begin{figure}[!ht]
\centering
\begin{minipage}{1.6in}
\centering
\includegraphics[height= 1.1in, width=1.5in ]{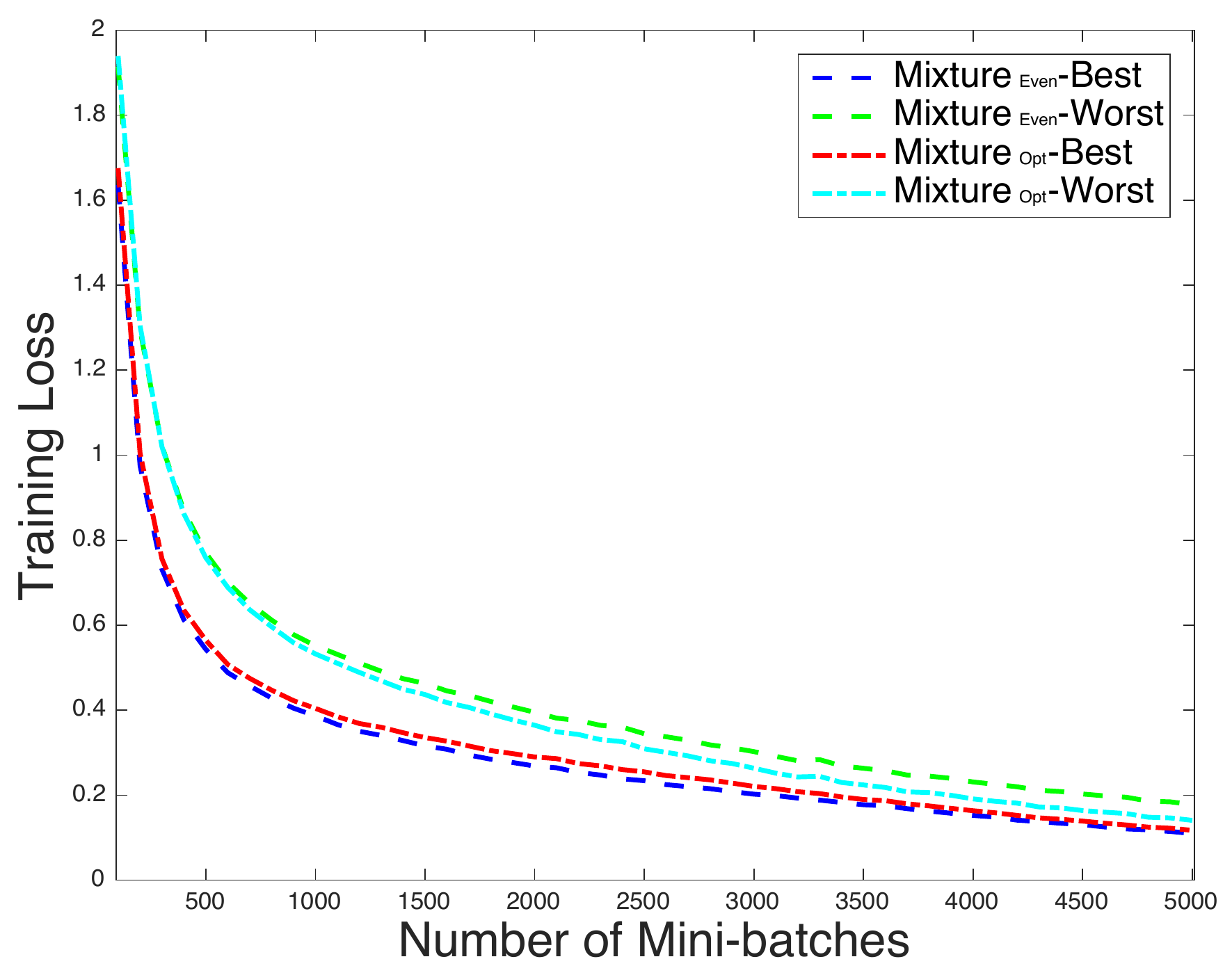}
\mbox{\footnotesize (a) $\sigma\in\{0, 4,8,12\}$ }
\end{minipage}
\begin{minipage}{1.6in}
\centering
\includegraphics[height= 1.1in, width=1.5in ]{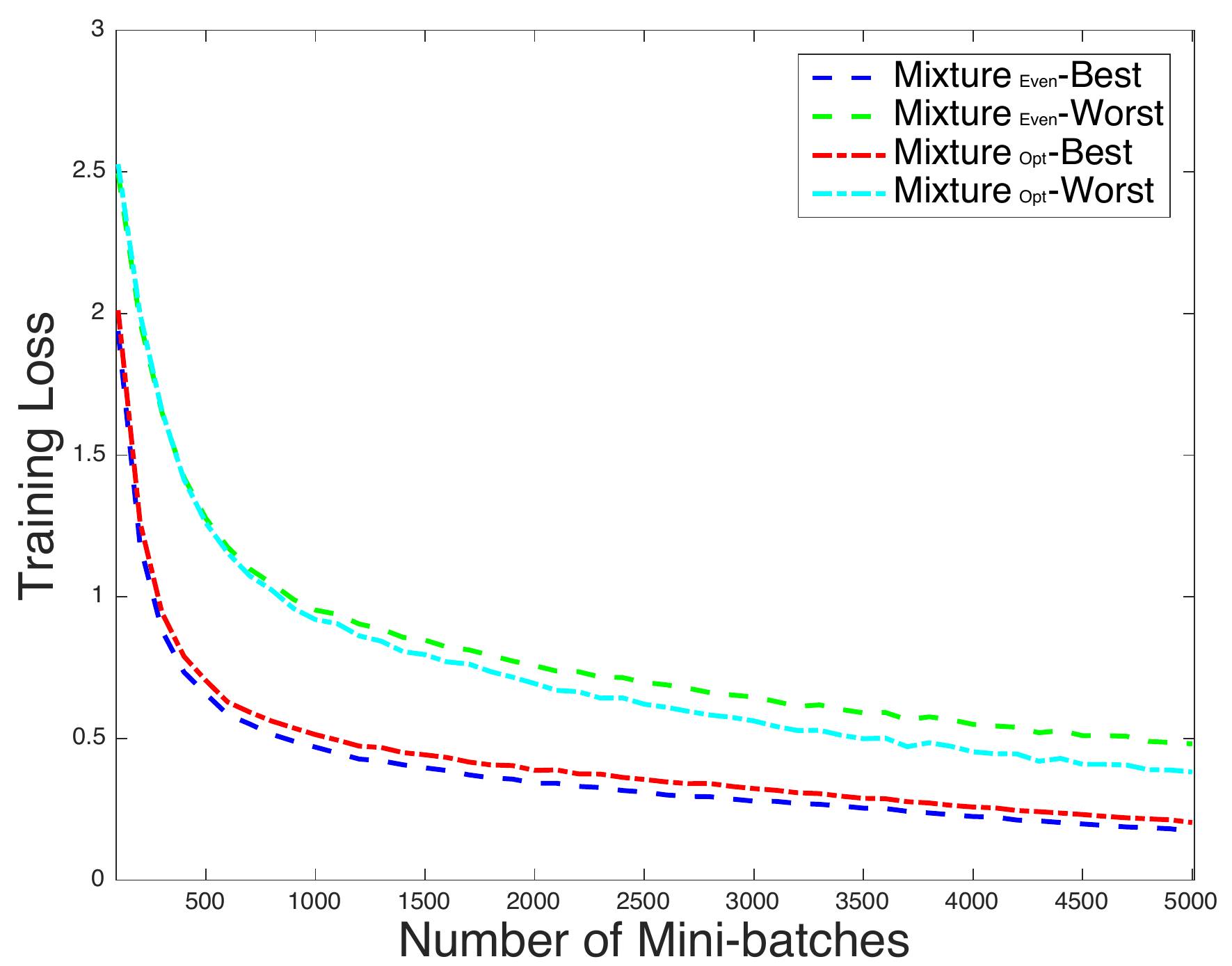}
\mbox{\footnotesize (b) $\sigma\in\{0, 10,20,30\}$}
\end{minipage}
\caption{\label{fig:noise}Illustration of best and worst training loss on ImageNet with Gaussian noise $\mathcal{N}(0,\sigma^2)$.}
\end{figure}

We adopt ResNet18~\cite{HeZRS16} as the base model in this experiment. It is initialized with the parameters learned from ILSVRC2012~\cite{ILSVRC15} and we set the learning rate as $\eta_w=0.005$ for fine-tuning. Considering the small size of data sets, we also include the method of \cite{ChenLSS17} in comparison and it is denoted as \textbf{Mixture}$_{\mathrm{Oracle}}$. Since the near-optimal oracle is infeasible for \textbf{Mixture}$_{\mathrm{Oracle}}$, we apply the model with $100$ SGD iterations instead as suggested in \cite{ChenLSS17}. The prior distribution in the regularizer is set to the uniform distribution.

\begin{figure*}[!ht]
\centering
\begin{minipage}{1.7in}
\centering
\includegraphics[height= 1.2in, width=1.6in ]{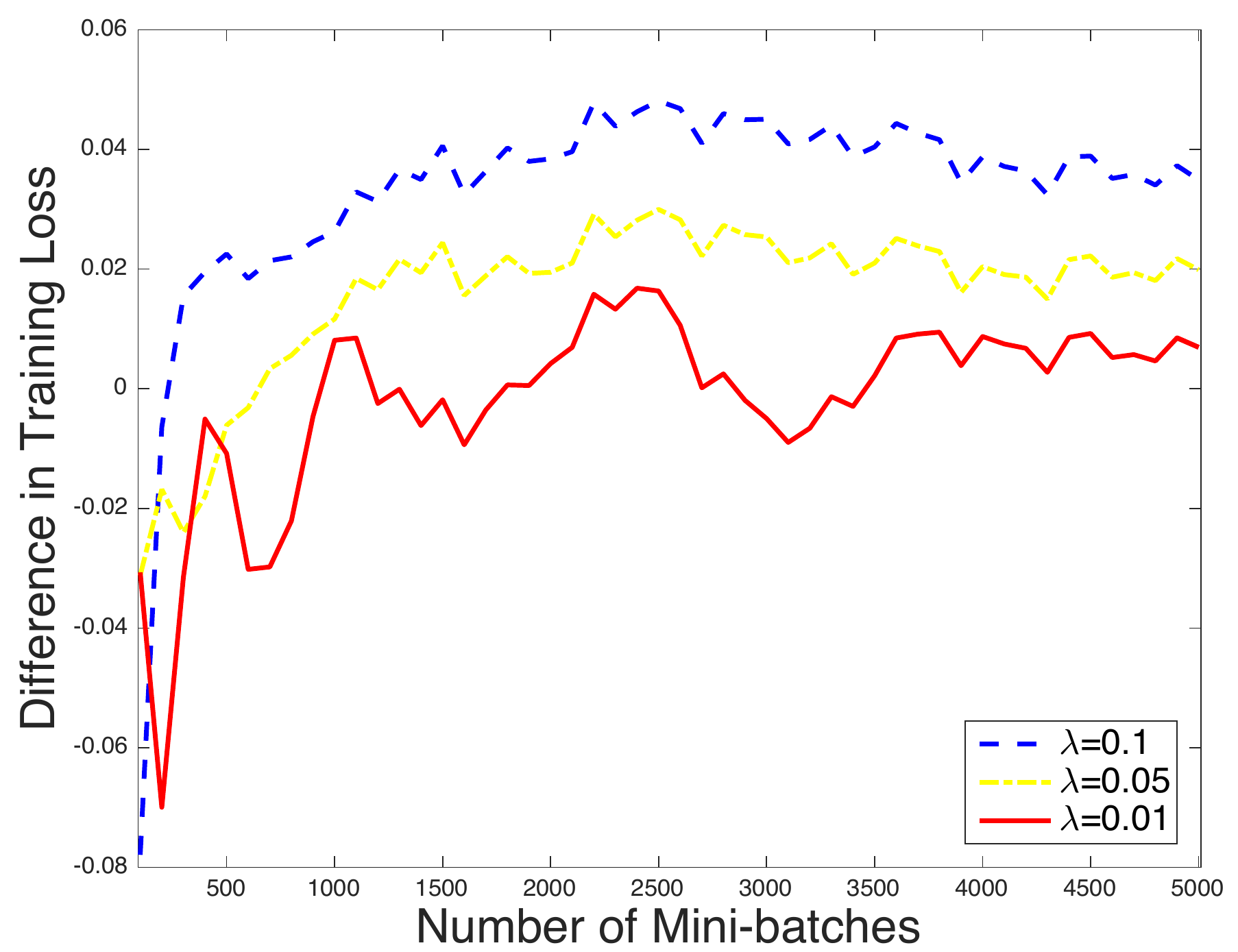}
\mbox{\footnotesize (a) $\D_{L_2}(\p||\q)$ }
\end{minipage}
\begin{minipage}{1.7in}
\centering
\includegraphics[height= 1.2in, width=1.6in ]{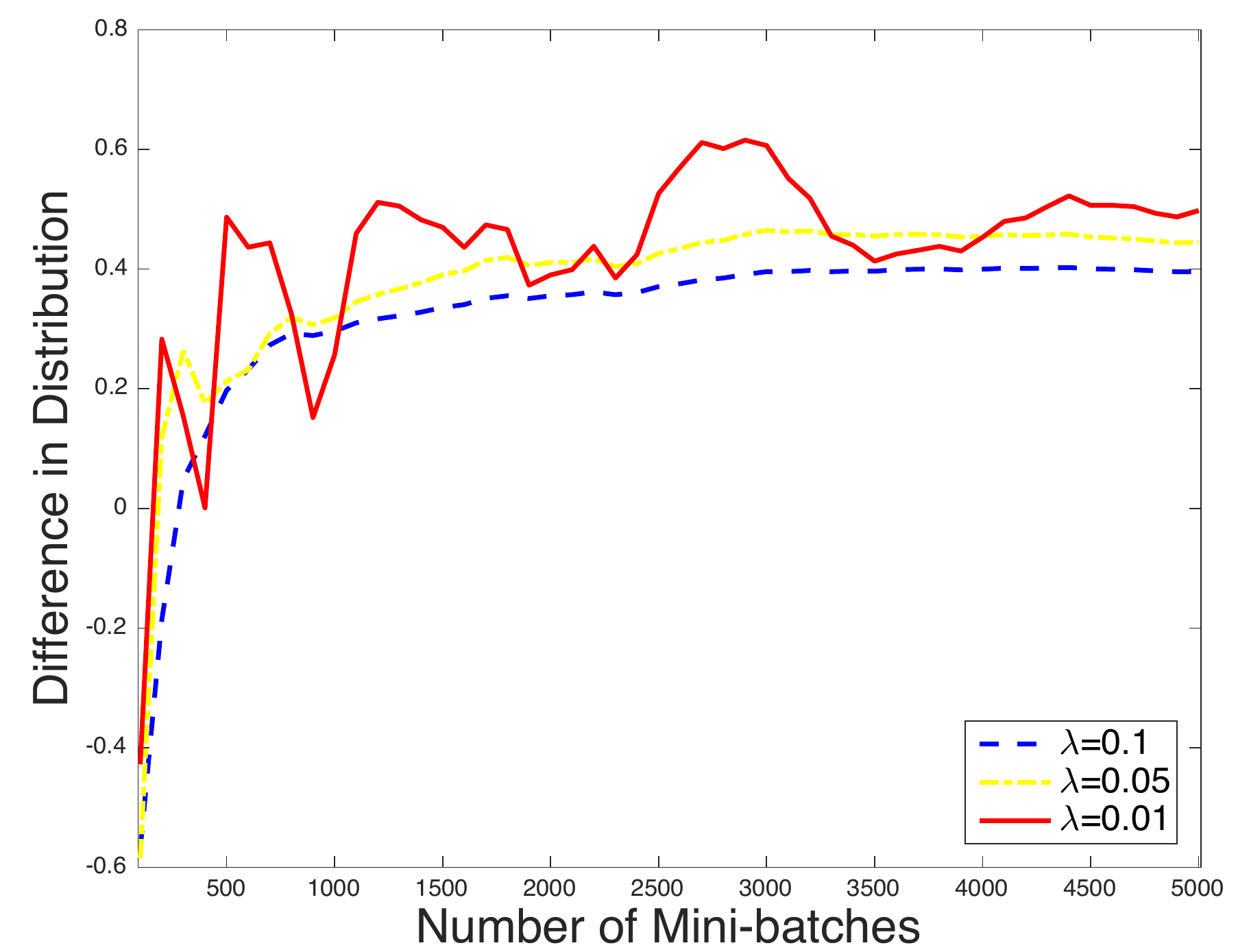}
\mbox{\footnotesize (b) $\D_{L_2}(\p||\q)$}
\end{minipage}
\begin{minipage}{1.7in}
\centering
\includegraphics[height= 1.2in, width=1.6in ]{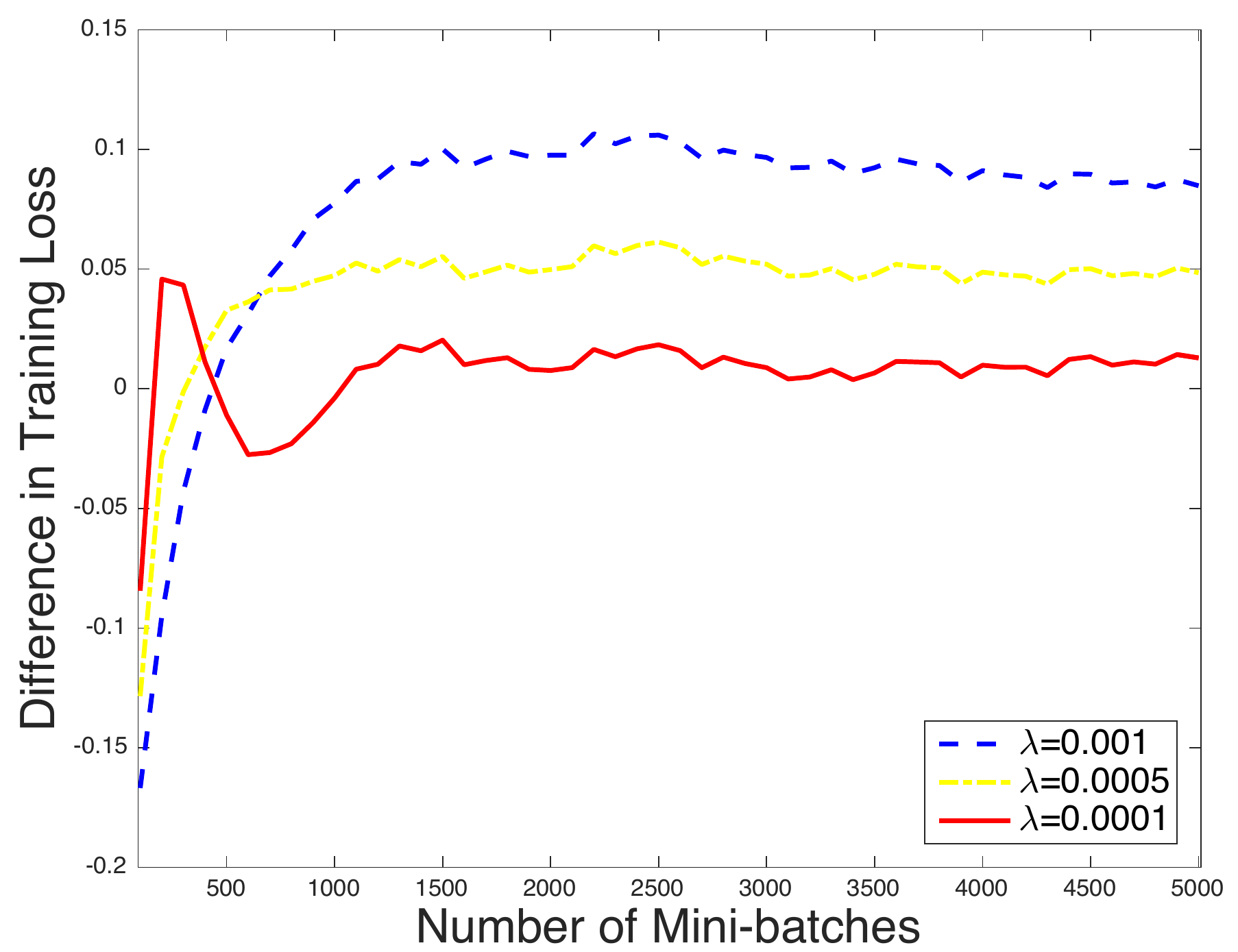}
\mbox{\footnotesize (c)  $\D_{\mathrm{OT}}(\p||\q)$}
\end{minipage}
\begin{minipage}{1.7in}
\centering
\includegraphics[height= 1.2in, width=1.6in ]{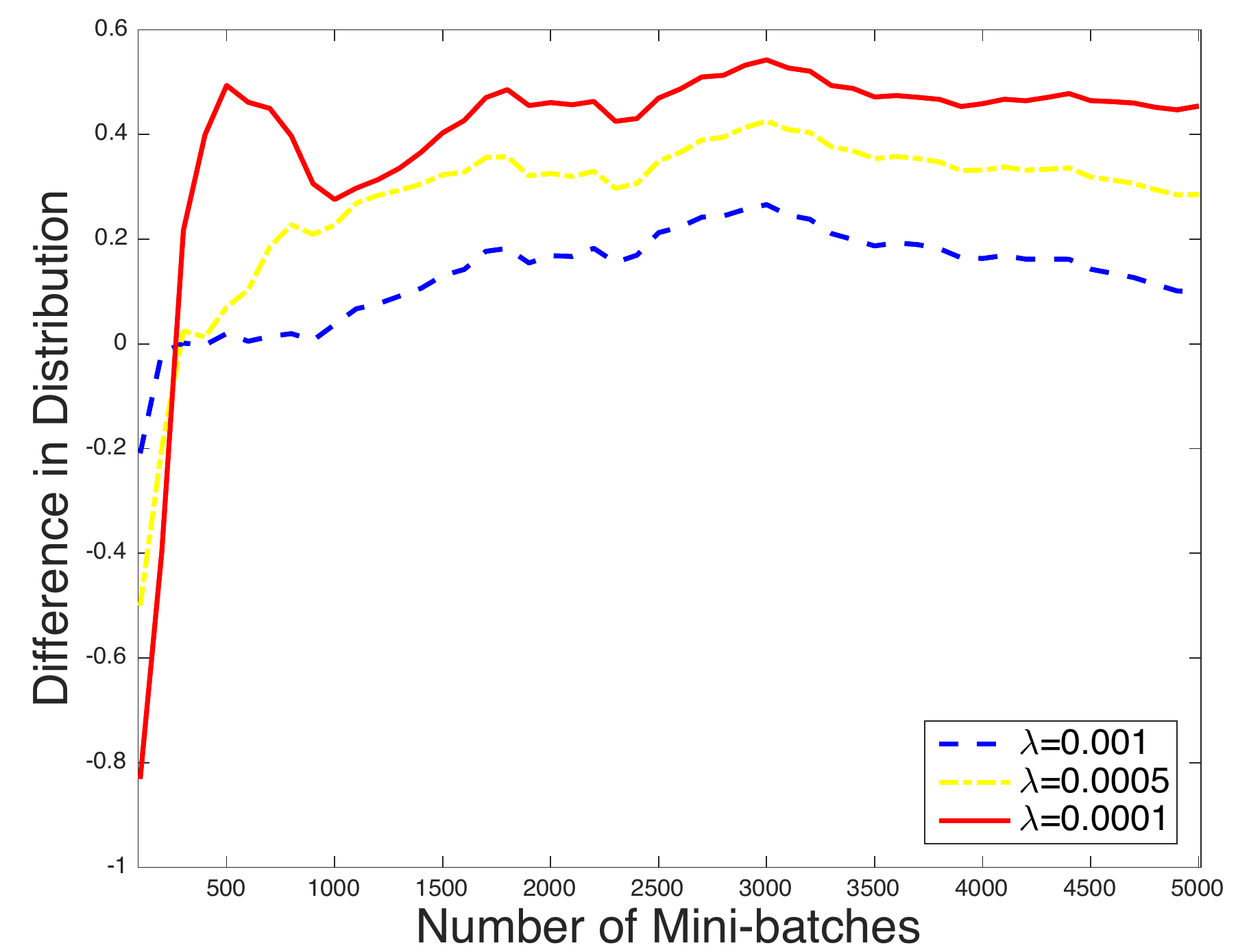}
\mbox{\footnotesize (d) $\D_{\mathrm{OT}}(\p||\q)$}
\end{minipage}
\caption{\label{fig:lambda}Illustration of the influence of the regularizer.}
\end{figure*}

Fig.~\ref{fig:tloss} summarizes the worst case training loss among multiple domains for the methods in the comparison. Since the performance of models learned from multiple domains is significantly better than those learned from an individual set, we illustrate the results in separate figures. Fig.~\ref{fig:tloss} (a) compares the proposed method to those with the individual data set. It is evident that the proposed method has the superior performance and learning with an individual domain cannot handle the data from other domains well. Fig.~\ref{fig:tloss} (b) shows the results of the methods learning with multiple data sets. First, we find that both Mixture$_{\mathrm{Oracle}}$ and Mixture$_{\mathrm{Opt}}$ can achieve the lower worst case loss than Mixture$_{\mathrm{Even}}$, which confirms the effectiveness of the robust optimization. Second, Mixture$_{\mathrm{Opt}}$ performs best among all of these methods and it demonstrates that the proposed method can optimize the performance over the adversarial distribution. To investigate the discrepancy between the performances on two domains, we illustrate the result in Fig.~\ref{fig:diffcd}. The discrepancy is measured by the difference between the empirical loss as $f_{\mathrm{ImageNet}}-f_{\mathrm{VGG}}$. We can find that $f_{\mathrm{ImageNet}}$ is smaller than $f_{\mathrm{VGG}}$ at the beginning but $f_{\mathrm{VGG}}$ decreases faster than $f_{\mathrm{ImageNet}}$. It is because the model is initialized with the parameters pre-trained on ImageNet. However, the task in VGG is easier than that in ImageNet, and $f_{\mathrm{VGG}}$ drops faster after a few iterations. Compared with the benchmark methods, the discrepancy from the proposed method is an order of magnitude better throughout the learning procedure. It verifies the robustness of Mixture$_{\mathrm{Opt}}$ and also shows that the proposed method can handle the drifting between multiple domains well. Finally, to compare the performance explicitly, we include the detailed results in Table~\ref{ta:fgvc}. Compared with the Mixture$_{\mathrm{Even}}$, we observe that Mixture$_{\mathrm{Opt}}$ can pay more attention to ImageNet than VGG and trade the performance between them.

\begin{table*}[!ht]
\small
\centering
\caption{Comparison on digits recognition.}\label{ta:digits}
\begin{tabular}{|l||l|l|l|l|l|l|l|l|}\hline
\multirow{2}{*}{Methods} &
\multicolumn{3}{l|}{MNIST} &\multicolumn{3}{l|}{SVHN}&\multirow{2}{*}{Acc$_{\mathrm{Tr}_\mathrm{w}}$}&\multirow{2}{*}{Acc$_{\mathrm{Te}_\mathrm{w}}$}\\
\cline{2-7}&Loss$_{\mathrm{Tr}}$&Acc$_{\mathrm{Tr}}$ &Acc$_{\mathrm{Te}} $&Loss$_{\mathrm{Tr}}$&Acc$_{\mathrm{Tr}}  $&Acc$_{\mathrm{Te}} $&&\\\hline
Individual$_{\mathrm{MNIST}}$&$0.001$&$100.00$&$98.81$&$4.01$&$30.80$&$29.58$&$30.80$&$29.58$\\\hline
Individual$_{\mathrm{SVHN}}$&$1.91$&$66.66$&$68.25$&$0.10$&$97.11$&$91.84$&$66.66$&$68.25$\\\hline
Mixture$_{\mathrm{Even}}$&$0.001$&$100.00$&$98.74$&$0.14$&$96.20$&$91.33$&$96.20$&$91.33$\\\hline
Mixture$_{\mathrm{Opt}}$&$0.03$&$99.03$&$98.13$&$0.11$&$97.05$&$92.14$&$\textbf{97.05}$&$\textbf{92.14}$\\\hline
\end{tabular}
\end{table*}

To further demonstrate that Mixture$_{\mathrm{Opt}}$ can trade the performance effectively, we conduct the experiments with noisy data. We simulate each individual domain by adding the random Gaussian noise from $\mathcal{N}(0,\sigma^2)$ to each pixel of the images from ImageNet pets. We vary the variance to generate the different domains and obtain two tasks where each has four domains with $\sigma\in\{0,4,8,12\}$ and $\sigma\in\{0,10,20,30\}$, respectively. Fig.~\ref{fig:noise} compares the gap between the best and worst performance on different domains for Mixture$_{\mathrm{Even}}$ and Mixture$_{\mathrm{Opt}}$. First, we can find that the proposed method improves the worst-case performance significantly while keeping the best performance almost the same. Besides, domains can achieve the similar performance for the simple task with variance in $\{0,4,8,12\}$. For the hard task that includes an extreme domain with noise from $\mathcal{N}(0,30^2)$, the best performance is not sacrificed much due to the appropriate regularizer in Mixture$_{\mathrm{Opt}}$.

After the comparison of performance, we illustrate the influence of the parameter $\lambda$ in Fig.~\ref{fig:lambda}. The parameter can be found in Eqn.~\ref{eq:problemr} and it constrains the distance of the adversarial distribution to the prior distribution. Besides the $L_2$ regularizer applied in Mixture$_{\mathrm{Opt}}$, we also include the results of the $\mathrm{OT}$ regularizer defined in Proposition~\ref{prop:ot} and the method is denoted as \textbf{Mixture}$_{\mathrm{OT}}$. Fig.~\ref{fig:lambda} (a) and (c) compare the discrepancy between the losses as in previous experiments. It is obvious that the smaller the $\lambda$, the smaller the gap between two domains. Fig.~\ref{fig:lambda} (b) and (d) summarize the drifting in a distribution, which is defined as $p_{\mathrm{ImageNet}}-p_{\mathrm{VGG}}$. Evidently, the learned adversarial distribution can switch adaptively according to the performance of the current model and the importance of multiple domains can be constrained well by setting $\lambda$ appropriately.

Finally, we compare the running time in Fig.~\ref{fig:rtime}. Due to the lightweight update for the adversarial distribution, Mixture$_{\mathrm{Opt}}$ and Mixture$_{\mathrm{OT}}$ have almost the same running time as Mixture$_{\mathrm{Even}}$. Mixture$_{\mathrm{Oracle}}$ has to enumerate the whole data set after each $100$ SGD iterations to update the current distribution, hence, its running time with only $50$ complete iterations is nearly $3$ times slower than the proposed method with $5,000$ iterations on these small data sets.

\subsection{Digits Recognition}
In this experiment, we examine the methods on the task of digits recognition, which is to identify 10 digits (i.e., $0$-$9$) from images. There are two benchmark data sets for the task: MNIST and SVHN. MNIST~\cite{lecun1998gradient} is collected for recognizing handwritten digits. It contains $60,000$ images for training and $10,000$ images for test. SVHN~\cite{netzer2011reading} is for identifying the house numbers from Google Street View images, which consists of $604,388$ training images and $26,032$ test images. Note that the examples in MNIST are $28\times 28$ gray images while those in SVHN are $32\times 32$ color images. To make the format consistent,  we resize images in MNIST to be $32\times 32$ and repeat the gray channel in RGB channels to generate the color images. Considering the task is more straightforward than pets categorization, we apply the AlexNet~\cite{KrizhevskySH12} as the base model in this experiment and set the learning rate as $\eta_w=0.01$. With a different deep model, we also demonstrate that the proposed framework can incorporate with various deep models.

Fig.~\ref{fig:tloss} (c) and (d) show the comparison of the worst case training loss and Table~\ref{ta:digits} summarizes the detailed results. We can observe the similar conclusion as the experiments on pets categorization. Mixture$_{\mathrm{Even}}$ can achieve good performance on these simple domains while the proposed method can further improve the worst case performance and provide a more reliable model for multiple domains.

\section{Conclusion}
\label{sec:conclusion}
In this work, we propose a framework to learn a robust model over multiple domains, which is essential for the service of cloud computing. The introduced algorithm can learn the model and the adversarial distribution simultaneously, for which we provide a theoretical guarantee on the convergence rate. The empirical study on real-world applications confirms that the proposed method can obtain a robust non-convex model. In the future, we plan to examine the performance of the method with more applications. Besides, extending the framework to multiple domains with partial overlapped labels is also important for real-world applications.

\section{ Acknowledgments}
We would like to thank Dr. Juhua Hu from University of Washington Tacoma and anonymous reviewers for their valuable suggestions that help to improve this work.

\bibliographystyle{aaai}
\bibliography{drl}

\section{Appendix}
\subsection{Proof of Lemma~1}
\begin{proof}
According to the updating criterion, we have
\begin{eqnarray}\label{eq:kl}
D_{KL}(\p||\p_{t+1}) - D_{KL}(\p||\p_{t}) =-\eta_p \p^\top \hat{\f}^t + \log(Z_t)
\end{eqnarray}
where $D_{KL}(\p||\q)$ denotes the KL-divergence between the distribution $\p$ and $\q$.
Note that for $a\in[0,1]$, we have
\[ \log(1-a(1-\exp(x)))\leq ax+x^2/8\]
Therefore
\begin{eqnarray*}
\sum_k\log(1-p_t^k(1-\exp(\eta_p \hat{f}_k^t)))\leq  \eta_p \p_t^\top \hat{\f}^t+\eta_p^2\|\hat{\f}^t\|_2^2/8
\end{eqnarray*}
Since $\hat{f}_k^t\geq 0$, we have $-p_t^k(1-\exp(\eta_p \hat{f}_k^t))\geq 0$ and 
\begin{align*} 
&\log(Z_t)=\log(1-\sum_k p_t^k(1-\exp(\eta_p\hat{f}_k^t)))\\
&\leq  \sum_k\log(1-p_t^k(1-\exp(\eta_p \hat{f}_k^t)))\leq \eta_p \p_t^\top \hat{\f}^t+\frac{\eta_p^2\|\hat{\f}^t\|_2^2}{8}
\end{align*}
Take it back to Eqn.~\ref{eq:kl} and we have
\begin{eqnarray}\label{eq:linearp}
(\p-\p_t)^\top \hat{\f}^t\leq \frac{D_{KL}(\p||\p_{t}) - D_{KL}(\p||\p_{t+1})}{\eta_p}+\frac{\eta_p\gamma^2}{8}
\end{eqnarray}
Therefore, for the arbitrary distribution $\p$, we have
\begin{align}\label{eq:p}
&\LL(\p,W_t) - \LL(\p_t,W_t) = (\p-\p_t)^\top \f(W_t)\nonumber\\
&= (\p-\p_t)^\top \hat{\f}^t+(\p-\p_t)^\top (\f-\hat{\f}^t)\nonumber\\
&\leq \frac{\eta_p\gamma^2}{8}+ \frac{D_{KL}(\p||\p_{t}) - D_{KL}(\p||\p_{t+1}) }{\eta_p} +(\p-\p_t)^\top (\f-\hat{\f}^t)
\end{align}

On the other hand, due to the convexity of the loss function, we have the inequality for the arbitrary model $W$ as
\begin{align}\label{eq:w}
&\LL(\p_t,W_{t})\leq \LL(\p_t,W)+\langle g_t, W_t-W\rangle\nonumber\\
&= \LL(\p_t,W)+\langle \hat{g}_t, W_{t}-W\rangle+\langle g_t-\hat{g}_t, W_t-W\rangle\nonumber\\
&\leq \LL(\p_t,W)+ \frac{\|W-W_t\|_F^2 -\|W-W_{t+1}\|_F^2  }{2\eta_w}\nonumber\\
&+\frac{\eta_w\sigma^2}{2}+\langle g_t-\hat{g}_t, W_t-W\rangle
\end{align}

Combine Eqn.~\ref{eq:p} and Eqn.~\ref{eq:w} and add $t$ from 1 to T
\begin{align*}
&\sum_t\LL(\p,W_t) - \LL(\p_t,W) \leq \frac{\log(K)}{\eta_p}+\frac{\|W-W_0\|_2^2}{2\eta_w}+\frac{T\eta_p\gamma^2}{8}\\
&+\frac{T\eta_w\sigma^2}{2}+\sum_t (\p-\p_t)^\top(\f-\hat{\f}^t)+\sum_t\langle g_t-\hat{g}_t,W_t-W\rangle
\end{align*}
where we use $D_{KL}(\p||\p_{0})\leq \log(K)$ with the fact that $\p_0$ is the uniform distribution.

Note that $\forall t$, we have $E[(\p-\p_t)^\top(\f-\hat{\f}^t)] = 0$ and $|(\p-\p_t)^\top(\f-\hat{\f}^t)|\leq \|f-\hat{f}^t\|_2\|p-p_t\|_2\leq 2\gamma$. According to the Hoeffding-Azuma inequality for Martingale difference sequence~\cite{cesa2006prediction}, with a probability $1-\delta$, we have
\[\sum_t (\p-\p_t)^\top(\f-\hat{\f}^t)\leq 2\sqrt{\gamma T\log(1/\delta)}\]

By taking the similar analysis, with a probability $1-\delta$, we have
\[\sum_t \langle g_t-\hat{g}_t, W_t-W\rangle\leq 2\sqrt{2\sigma R T\log(1/\delta)}\]

Therefore, when setting $\eta_w = \frac{R}{\sigma\sqrt{T}}$ and $\eta_p = \frac{2\sqrt{2\log(K)}}{\gamma\sqrt{T}}$, with a probability $1-\delta$, we have
\begin{align*}
&\sum_t\LL(\p,W_t) - \LL(\p_t,W) \leq c_1\sqrt{T}+2c_2\sqrt{T\log(2/\delta)}
\end{align*}
where $c_1$ and $c_2$ are
\[c_1 = \gamma\sqrt{\frac{\log(K)}{2}}+\sigma R;\quad c_2 = \sqrt{\gamma}+\sqrt{2\sigma R}\]

Due to the convexity of $\LL(\cdot,\cdot)$ in $W$ and concavity in $\p$, with a probability $1-\delta$, we have
\begin{eqnarray*}
\LL(\p,\overline{W}) -\LL(\bar{\p},W)&\leq& \frac{1}{T} \sum_t\LL(\p,W_t) - \LL(\p_t,W) \\
&\leq& \frac{c_1}{\sqrt{T}}+\frac{2c_2\sqrt{\log(2/\delta)}}{\sqrt{T}}
\end{eqnarray*}
We finish the proof by taking the desired $(\p,W)$ into the inequality.
\end{proof}

\subsection{Proof of Lemma~2}
\begin{proof}
We first present some necessary definitions. 
\begin{definition}
A function $F$ is called $L$-smoothness w.r.t. a norm $\|\cdot\|$ if there is a constant $L$ such that for any $W$ and $W'$, it holds that
\[F(W')\leq F(W)+\langle \nabla F(W),W'-W\rangle+\frac{L}{2}\|W'-W\|^2 \]
\end{definition}

\begin{definition}
A function $F$ is called $\lambda$-strongly convex w.r.t. a norm $\|\cdot\|$ if there is a constant $\lambda$ such that for any $W$ and $W'$, it holds that
\[F(W')\geq F(W)+\langle \nabla F(W),W'-W\rangle+\frac{\lambda}{2}\|W'-W\|^2 \]
\end{definition}

According to the $L$-smoothness of the loss function, we have
\begin{align*}
&E[\LL(\p_t,W_{t+1})]\leq E[\LL(\p_t,W_t)+\langle g_t,W_{t+1}-W_t\rangle\\
&+\frac{L}{2}\|W_{t+1}-W\|_F^2]\\
&\leq E[\LL(\p_t,W_t)-\eta_w\langle g_t,\hat{g}_t\rangle+\frac{L\eta_w^2}{2}\|\hat{g}_t\|_F^2]\\
&\leq E[\LL(\p_t,W_t)]-\eta_wE[\|g_t\|_F^2]+\frac{L\eta_w^2\sigma^2}{2}
\end{align*}
So we have
\begin{align}\label{eq:nonconv}
&E[\|g_t\|_F^2]\leq \frac{E[\LL(\p_t,W_t)-\LL(\p_t,W_{t+1})]}{\eta_w}+\frac{L\eta_w \sigma^2}{2}\nonumber\\
&= \frac{E[\LL(\p_t,W_t)-\LL(\p_{t+1},W_{t+1})]}{\eta_w}\nonumber\\
&+\frac{E[\LL(\p_{t+1},W_{t+1})-\LL(\p_{t},W_{t+1})]}{\eta_w}+\frac{L\eta_w \sigma^2}{2}
\end{align}


Now we try to bound the difference between $\LL(\p_{t+1},W_{t+1})$ and $\LL(\p_{t},W_{t+1})$
\begin{align}
&E[\LL(\p_{t+1},W_{t+1})-\LL(\p_{t},W_{t+1})]=E[(\p_{t+1}-\p_t)^\top \f(W_{t+1})]\nonumber\\
&\leq E[\|\p_{t+1}-\p_t\|_2\|\f(W_{t+1})\|_2] \leq \gamma E[\|\p_{t+1}-\p_t\|_1]\nonumber\\
&\leq \gamma E[\sqrt{2D_{KL}(\p_{t}||\p_{t+1})}]\label{eq:pinsker}\\
&\leq \eta_p\gamma^2/2\label{eq:klbound}
\end{align}
Eqn.~\ref{eq:pinsker} is from the Pinsker's inequality and Eqn.~\ref{eq:klbound} is from the inequality in Eq.\ref{eq:linearp} by letting $\p = \p_t$.

Adding Eqn.~\ref{eq:nonconv} from $1$ to $T$ with Eqn.~\ref{eq:klbound}, we have
\begin{eqnarray*}
\sum_tE[\|g_t\|_F^2]\leq \frac{\LL(\p_0,W_0)}{\eta_w}+\frac{\eta_pT\gamma^2}{2\eta_w}+\frac{TL\eta_w \sigma^2}{2}
\end{eqnarray*}
On the other hand, with the similar analysis in Eqn.~\ref{eq:p}, we have
\[\sum_t E[\LL(\p,W_t)] - E[\LL(\p_t,W_t)]\leq \frac{\log(K)}{\eta_p}+\frac{T\eta_p\gamma^2}{8}\]
\end{proof}

\subsection{Proof of Theorem~2}
\begin{proof}

Since $\LL(\p,W)$ is $\lambda$-strongly concave in $\p$, we have
\begin{align*}
&E[\LL(\p ,W_t)-\LL(\p_t,W_t)]\leq E[(\p-\p_t)^\top h_t-\frac{\lambda}{2}\|\p-\p_t\|_2^2]\\
&= E[(\p-\p_t)^\top \hat{h}_t-\frac{\lambda}{2}\|\p-\p_t\|_2^2]\\
&\leq \frac{\eta_p^t \mu^2}{2}+ \frac{E[\|\p-\p_t\|_2^2]-E[\|\p-\p_{t+1}\|_2^2]}{2\eta_p^t}-\frac{\lambda}{2}E[\|\p-\p_t\|_2^2]
\end{align*}

Taking $\eta_p^t = \frac{1}{\lambda t}$ and add the equation from $1$ to $T$, we have
\begin{eqnarray*}
&&E[\sum_t\LL(\p,W_t)-\LL(\p_t,W_t)]\leq \frac{\mu^2}{2\lambda}\sum_t \frac{1}{t}\leq \frac{\mu^2\log(T)}{\lambda}
\end{eqnarray*}

On the other hand, we have
\begin{align*}
&E[\LL(\p_{t+1},W_{t+1}) - \LL(\p_{t},W_{t+1})]\\
&\leq E[(\p_{t+1}-\p_t)^\top\nabla \LL_{\p_t}(\p_t,W_{t+1})]\\
&=E[(\p_{t+1}-\p_t)^\top \hat{h}^{t+1}]+E[\lambda\|\p_{t+1}-\p_t\|_2^2]\\
&\leq \eta_p^t\mu^2 +\lambda(\eta_p^t)^2\mu^2
\end{align*}

Take it back to Eqn.~\ref{eq:nonconv} and add $t$ from $1$ to $T$, then we have
\begin{align*}
&\sum_tE[\|g_t\|_F^2]\leq \frac{\LL(\p_0,W_0)}{\eta_w}+\frac{\sum_t\eta_p^t\mu^2+\lambda(\eta_p^t)^2\mu^2}{\eta_w}+\frac{TL\eta_w \sigma^2}{2}\\
&\leq \frac{\LL(\p_0,W_0)}{\eta_w}+\frac{(\pi^2/6+2\log(T))\mu^2}{\lambda \eta_w}+\frac{TL\eta_w \sigma^2}{2}
\end{align*}
We finish the proof by letting $\eta_w = \frac{2\mu\sqrt{\log(T)}}{\sigma\sqrt{\lambda L T}}$.
\end{proof}

\subsection{Proof of Theorem~4}
\begin{proof}
According to the $L$-smoothness of the loss function, we have
\begin{align*}
&E[\F(W_{t+1})]\leq E[\F(W_t)+\langle \nabla \F(W_t),W_{t+1}-W_t\rangle\\
&+\frac{L}{2}\|W_{t+1}-W\|_F^2]\\
&\leq E[\F(W_t)-\eta_w\langle \nabla \F(W_t),\hat{g}_t\rangle+\frac{L\eta_w^2}{2}\|\hat{g}_t\|_F^2]\\
&\leq E[\F(W_t)]-\eta_wE[\|\nabla \F(W_t)\|_F^2] \\
&+ \eta_w\langle\nabla \F(W_t),\nabla \F(W_t)-\hat{g}_t\rangle+\frac{L\eta_w^2\sigma^2}{2}\\
&\leq E[\F(W_t)]-\eta_wE[\|\nabla \F(W_t)\|_F^2] \\
&+ \eta_w\sigma^2\|\p_t^* - \p_t\|_1+\frac{L\eta_w^2\sigma^2}{2}
\end{align*}
So we have
\begin{align*}
&E[\|\nabla \F(W_t)\|_F^2]\leq  \frac{E[\F(W_t)-\F(W_{t+1})]}{\eta_w}+\sigma^2\xi_t+\frac{L\eta_w \sigma^2}{2}
\end{align*}
Adding inequalities from $1$ to $T$, we have
\[\sum_t E[\|\nabla \F(W_t)\|_F^2] \leq \frac{\F(W_0)}{\eta_w}+2\sigma^2\sqrt{T}+\frac{TL\eta_w \sigma^2}{2}\]
We complete the proof by setting $\eta_w = \frac{\sqrt{2}}{\sqrt{TL}\sigma}$.
\end{proof}

\begin{figure}[!ht]
\centering
\includegraphics[width=0.3\textwidth]{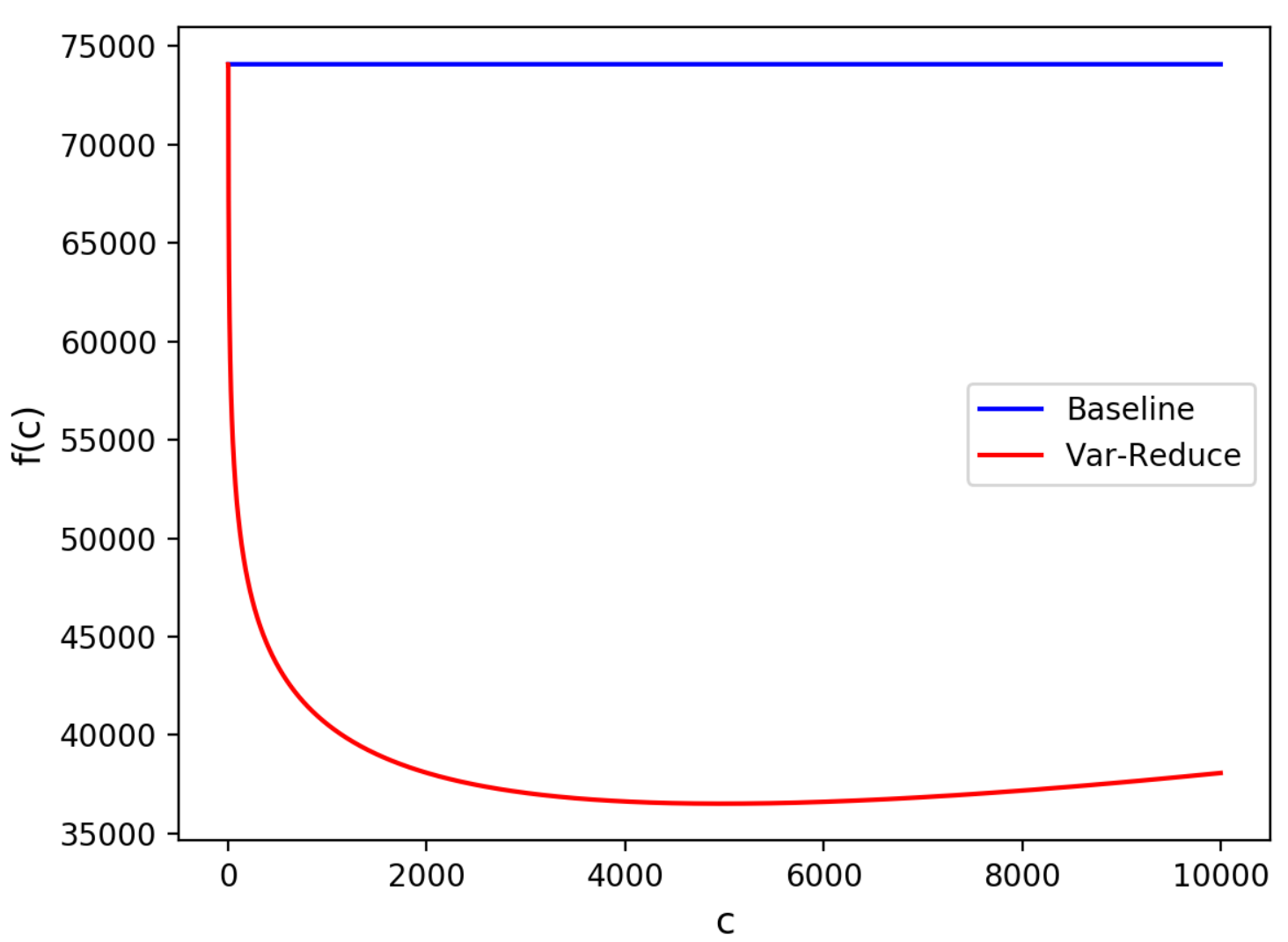}
\caption{Illustration of the improvement from the step-size.}\label{fig:illu}
\end{figure}

\subsection{Proof of Proposition~1}
By taking the closed-form solution for $P$, we have
\begin{align*}
&\D_{\mathrm{OT}}(\p||\q) = \max_{\alpha,\beta}-\sum_{i,j}\frac{1}{\lambda}\exp(-1-\lambda(m_{ij}+\alpha_i+\beta_j))\\
&-\alpha^\top \p-\beta^\top \q
\end{align*}

Given two distributions $\p_x$ and $\p_y$ and let $t\in[0,1]$, then we have
\begin{align*}
&D_{\mathrm{OT}}(t\p_x+(1-t)\p_y||\q) \\
&=  \max_{\alpha,\beta}-\sum_{i,j}\frac{1}{\lambda}\exp(-1-\lambda(m_{ij}-\alpha_i-\beta_j))\\
&-\alpha^\top (t\p_x+(1-t)\p_y)-\beta^\top \q\\
& =  \max_{\alpha,\beta}-t\sum_{i,j}\frac{1}{\lambda}\exp(-1-\lambda(m_{ij}-\alpha_i-\beta_j))\\
&-(1-t)\sum_{i,j}\frac{1}{\lambda}\exp(-1-\lambda(m_{ij}-\alpha_i-\beta_j))\\
&-t\alpha^\top\p_x - (1-t)\alpha^\top \p_y-t\beta^\top \q-(1-t)\beta^\top \q\\
&\leq t D_{\mathrm{OT}}(\p_x||\q)+(1-t) D_{\mathrm{OT}}(\p_y||\q)
\end{align*}

Therefore $\D_{\mathrm{OT}}(\p||\q)$ is convex in $\p$ and 
\[\nabla \D_{\mathrm{OT}}(\p||\q) = \alpha^*\]
where $\alpha^*$ is the optimal solution for the maximizing problem with the prior distribution $\q$. It can be obtained by Sinkhorn-Knopp's fixed point iteration efficiently~\cite{Cuturi13}.

\subsection{Proof of Corollary~1}
\begin{proof}
First, we show that RHS of Theorem~3 is concave in $c$.
Let
\[f(c) = \lambda c+ \frac{\mu^2}{2\lambda}\ln(\frac{T}{c}+1)+\frac{\mu^2}{2\lambda} \]
It is a convex function when $c>0$, because
\[f''(c) = \frac{\mu^2(T^2+2Tc)}{2\lambda(Tc+c)^2}\geq 0\]
Therefore $-f(c)$ is concave and the optimal value can be obtained by setting the gradient to zero as
\[f'(c) = \lambda - \frac{\mu^2T}{2\lambda(Tc+c^2)}=0\]
$c$ has the closed-form solution as
\[c = \frac{\sqrt{T^2+\frac{2\mu^2T}{\lambda^2}}-T}{2} = \frac{\mu^2}{\lambda^2(1+\sqrt{1+\frac{2\mu^2}{\lambda^2T}})}\]
\end{proof}
To illustrate the influence of $c$, we show an example when $T=1e6$, $\mu=1e2$ and $\lambda=1$ in Fig.~\ref{fig:illu}. First, we define the regret of the algorithm as
 \[\mathrm{Regret}=\max_{\p\in\Delta}\sum_t\hat{\LL}(\p,W_t) - \sum_t\hat{\LL}(\p_t,W_t))\]
The baseline is the regret of the conventional step-size $\eta_p^t = \frac{1}{\lambda t}$, which is $\frac{\mu^2}{2\lambda}(\ln(T)+1)$. The regret of the proposed step-size is denoted by the red line and it shows the regret can be significantly reduced when setting the constant $c$ to optimum.

\end{document}